\pdfoutput=1

\documentclass[11pt]{article}

\usepackage[final]{acl}

\usepackage{times}
\usepackage{latexsym}

\usepackage[T1]{fontenc}

\usepackage[utf8]{inputenc}

\usepackage{microtype}

\usepackage{inconsolata}

\usepackage{graphicx}

\usepackage{amsmath,amsfonts,bm}

\def\eqref#1{equation~\ref{#1}}

\def\1{\bm{1}}

\DeclareMathAlphabet{\mathsfit}{\encodingdefault}{\sfdefault}{m}{sl}
\SetMathAlphabet{\mathsfit}{bold}{\encodingdefault}{\sfdefault}{bx}{n}

\usepackage{hyperref}
\usepackage{url}
\usepackage{booktabs}
\usepackage{multicol}
\usepackage{multirow}
\usepackage{amsthm}
\usepackage{multirow}
\usepackage{wrapfig}
\usepackage{float}
\usepackage{listings}
\usepackage{tcolorbox}
\usepackage{rotating}
\usepackage{amsthm}
\usepackage{tcolorbox}  %
\usepackage{lipsum}     %
\usepackage{titlesec}   %
\usepackage{hyperref}   %
\usepackage{etoc}
\usepackage{titletoc}

\newtheorem{lemma}{Lemma}
\theoremstyle{definition}
\newtheorem{definition}{Definition}
\theoremstyle{remark}
\newtheorem{remark}{Remark}
\theoremstyle{theorem}
\newtheorem{theorem}{Theorem}
\newtheorem{hypothesis}{Hypothesis}
\usepackage{colortbl}
\usepackage{scrextend}

\newcommand{\revise}[1]{{\color{black} #1}}

\title{QAEncoder: Towards Aligned Representation Learning \\ in Question Answering Systems}

\author{Zhengren Wang$^{1,2\dagger}$, Qinhan Yu$^{1\dagger}$, Shida Wei$^{1}$
\\  {\bf Zhiyu Li$^{2*}$, Feiyu Xiong$^{2}$, Xiaoxing Wang$^{2}$, Simin Niu$^{2}$, Hao Liang$^{1,3}$, Wentao Zhang$^{1,2,3*}$ }\\
$^1$Peking University~$^2$Institute for Advanced Algorithms Research~$^3$Zhongguancun Academy\\ 
\texttt{wzr@stu.pku.edu.cn} ~~~ \texttt{\{lizy,xiongfy\}@iaar.ac.cn} ~~~ \texttt{wentao.zhang@pku.edu.cn} 
}

\begin{document}

\maketitle
\begingroup
\deffootnote[1.5em]{1.5em}{1em}{}
\renewcommand\thefootnote{}\footnote{
$\dagger$ Equal contribution;  * Corresponding author.
}
\addtocounter{footnote}{-1}
\endgroup

\begin{abstract}
Modern QA systems entail retrieval-augmented generation (RAG) for accurate and trustworthy responses. However, the inherent gap between user queries and relevant documents hinders precise matching. We introduce QAEncoder, a training-free approach to bridge this gap. Specifically, QAEncoder estimates the expectation of potential queries in the embedding space as a robust surrogate for the document embedding, and attaches document fingerprints to effectively distinguish these embeddings. 
Extensive experiments across diverse datasets, languages, and embedding models confirmed QAEncoder's alignment capability, which offers a simple-yet-effective solution with zero additional index storage, retrieval latency, training costs, or catastrophic forgetting and hallucination issues. The repository is publicly available at \url{https://github.com/IAAR-Shanghai/QAEncoder}.

\end{abstract}

\section{Introduction}
\begin{quote}
    \textit{``What I cannot create, I do not understand."} 
    \qquad\qquad\;\;\, --- Richard Feynman
\end{quote}
Question Answering (QA) systems aim to generate accurate responses to user queries with applications in customer service \citep{xu2024retrieval}, search engine \citep{ojokoh2018review}, healthcare \citep{guo2022medical} and education \citep{levonian2023retrieval}.
Modern QA systems leverage large language models (LLMs) such as ChatGPT \citep{achiam2023gpt}, supplemented with retrieval-augmented generation (RAG) to address issues of outdated or hallucinatory information, especially for rapidly evolving knowledge bases \citep{{2020RAG,hallucination},gupta2024rag}. The efficacy of RAG hinges on its retrieval module for identifying relevant documents from a vast corpus. Dense retrievers \citep{2020RAG,hofstatter2021efficiently}, contrasted with keyword-matching-based sparse retrievers \citep{TF-IDF,BM25}, have enabled more precise retrieval by mapping queries and documents into a shared vector space.
Despite advancements, a significant challenge that persists is bridging the gap between user queries and documents across lexical, syntactic, semantic, and content dimensions \citep{zheng2020bert,Doc2query,docT5query}, termed \textit{the document-query gap}. Three main approaches have emerged to address this challenge: training-based, document-centric and query-centric alignment. 

Training-based approaches \citep{dong,learning,w-etal-2023-query,Mafin,tabular} directly train embedding models with QA datasets to close the representation of relevant queries and documents. But for out-of-domain or multi-domain adaption settings, they struggle to fully generalize \citep{generalizability} and face catastrophic forgetting issues. 
Document-centric approaches \citep{Query2doc,HyDE,QA-RAG} generate pseudo-documents for user queries via LLMs, which are then used as retrieval queries; however, this inference process is costly, time-consuming, and prone to hallucinations \citep{Query2doc}.

In contrast, query-centric methods \citep{Doc2query} generate and index potential queries to better match user queries. Yet, existing methods mainly focus on sparse retrievers and have not fully leveraged the potential of dense retrievers \citep{docT5query,mallia2021learning}. The naive attempt is to directly store predicted QA pairs into a vector database, but with evident drawbacks like expanded index size, longer retrieval latency, and limited query handling.  Therefore, integrating query-centric methods with dense retrievers remains challenging and heavily under-explored.

\begin{figure*}[tb!]
    \centering
    \includegraphics[width=\linewidth]{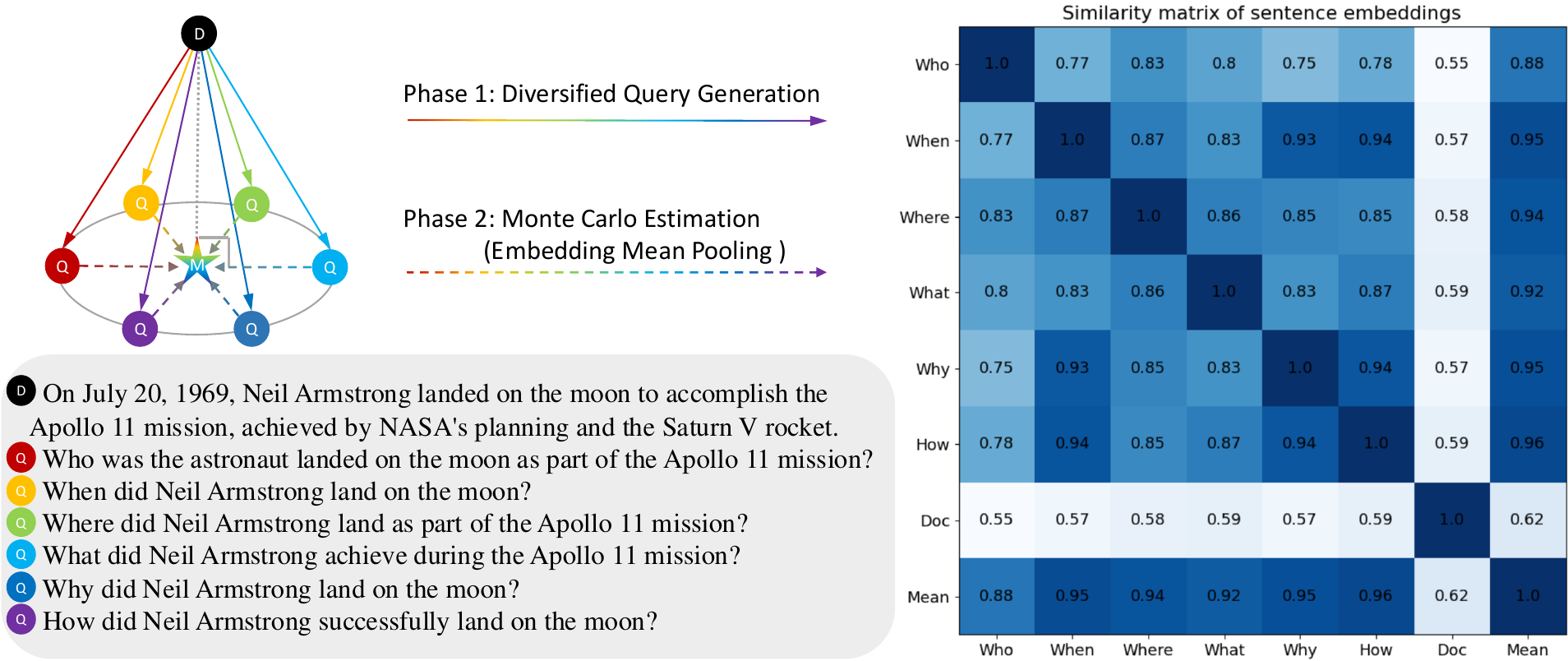}
    \caption{
    Illustration of QAEncoder's alignment process. Left: Solid lines represent diversified query generation, while dashed lines indicate Monte Carlo estimation.
    Right: 
    The heatmap depicts the similarity scores among the embeddings of different queries, the document, and the mean estimation. 
    Compared to the document itself, the mean estimation is significantly better aligned with different queries, i.e. a robust surrogate for the document embedding.
    }
    \label{fig:demo}
\end{figure*}

\paragraph{Motivation} Inspired by Feynman's philosophy of learning, in QA systems, effective information retrieval extends beyond mere storage and involves creative processes like query formulation and summarization \citep{reyes2021feynman}. For instance, the well-established 5W1H framework (Who, What, When, Where, Why, How) help systematically deconstruct information and actively foster a deeper understanding \citep{5W1H}. In this paper, we continue the research line of query-centric methods, and propose \textit{QAEncoder} as a pioneering work.

As demonstrated in Fig. \ref{fig:demo}, our method initially generates diversified queries (e.g. 5W1H), and then estimates the cluster center of potential queries by the Monte Carlo method. The similarity matrix reveals that, for any query, the mean-query similarity is significantly higher than both document-query and other query-query similarities. Hence, we advocate using the cluster center as a surrogate for the document embedding, which bridges the document-query gap robustly without extra index size and retrieval latency. Meanwhile, we propose the conical distribution hypothesis for more theoretical grounding, which intuitively visualizes the document-query gap and its alignment by abstracting geometric properties of the embedding space. 

Despite these advantages, this basic proposal encounters a critical challenge. While enhancing similarity with user queries, it simultaneously reduces the distinguishability between document representations, as they all become query-like. To address this side effect, we further propose \textit{document fingerprint strategies} to reintroduce unique document identities into representations and enable state-of-the-art performance.

\paragraph{Contributions} The contributions are threefold:
\begin{itemize}
    \item \textbf{Methodological Innovations.}
    We pioneer to bridge the document-query gap in dense retrievers from the query-centric perspective. 
    Our method, QAEncoder, not only avoids extra index storage, retrieval latency, training cost and hallucination, but also guarantees diversified query handling and robust generalization. We also propose document fingerprints to address the side effect on distinguishability and achieve state-of-the-art performance.
    \item \textbf{Theoretical Discovery.}
    We formulate the conical distribution hypothesis, and validate it through empirical analysis. This hypothesis provides not only deeper insights into the geometry of semantic space, but also theoretical foundations for the alignment process.
    \item \textbf{Practical Applications.}
    QAEncoder is a simple-yet-effective plugin, which seamlessly integrates with existing RAG architectures and training-based methods.
    This integration significantly boosts system performance with minor modifications required.
\end{itemize}

\section{Related Work}

\paragraph{Retrieval-augmented QA systems.}
Retrieval-augmented generation significantly improves large language models in QA systems by incorporating a retrieval module that fetches relevant information from external knowledge sources \citep{EaE, REALM, FID, survey}. Retrieval models have evolved from early sparse retrievers, such as TF-IDF \citep{TF-IDF} and BM25 \citep{BM25}, which rely on word statistics and inverted indices, to dense retrieval strategies \citep{2020RAG} that utilize neural representations for enhanced semantic matching. Advanced methods, such as Self-RAG \citep{Self-RAG} which determines if additional information is required and evaluates the relevance of retrieved content, and RAG-end2end \citep{RAG-end2end} that jointly trains the retriever and generator, represent significant developments in this area. However, these methods still ignore the document-query gap.

\paragraph{Training-based alignment.}
Training-based approaches bridge the document-query gap generally by contrastive learning \citep{xiongapproximate,qu2021rocketqa} or knowledge distillation \citep{Mafin, tabular}. For instance, \citet{dong} showed parameter sharing of the query encoder and the document encoder improves overall performance by projecting queries and documents into shared space. 
Dual-Cross-Encoder \citep{learning} and Query-as-context \citep{w-etal-2023-query} train embedding models from scratch with paired document-query samples. GPL \citep{wang2021gpl}, CAI \citep{iida2022unsupervised} and AugTriever \citep{meng2022augtriever} conduct domain adaptation with pseudo-queries.
However, training-based methods usually face generalization difficulty \citep{generalizability}, and catastrophic forgetting issues \citep{database,saunders2022domain}.

\paragraph{Document-centric alignment.}
Document-centric methods, such as HyDE \citep{HyDE} and Query2doc \citep{Query2doc}, dynamically transform user queries into pseudo-documents using LLMs for both sparse and dense retrievers. QA-RAG \citep{QA-RAG} advances by implementing a two-way retrieval mechanism that utilizes both user query and pseudo-documents for respective retrieval and ranking. 
However, their effectiveness is highly dependent on the quality of pseudo-documents, which are prone to hallucinations, especially for the latest information. Furthermore, during inference, invoking LLMs for user queries imposes extra computational cost and latency, leading to degraded user experience.

\paragraph{Query-centric alignment.}
The seminal work, Doc2Query \citep{Doc2query}, focuses on the vocabulary mismatch problem for sparse retrievers by expanding the document with keywords in predicted queries. The improved version, DocT5Query \citep{docT5query}, trains a T5 model to predict queries, but remains confined to sparse retrievers. There are also proposals to directly store question-answer pairs for retrieving and ranking, such as RePAQ \citep{lewis2021paq}, QUADRo and QR \citep{campese2023quadro,campese2024pre}. However, these attempts suffer from expanded index size, longer retrieval latency, and limited query handling. Hence, the query-centric alignment for dense retrievers remains challenging and largely unexplored.

\begin{figure*}[tb!]
    \centering
    \includegraphics[width=1\linewidth]{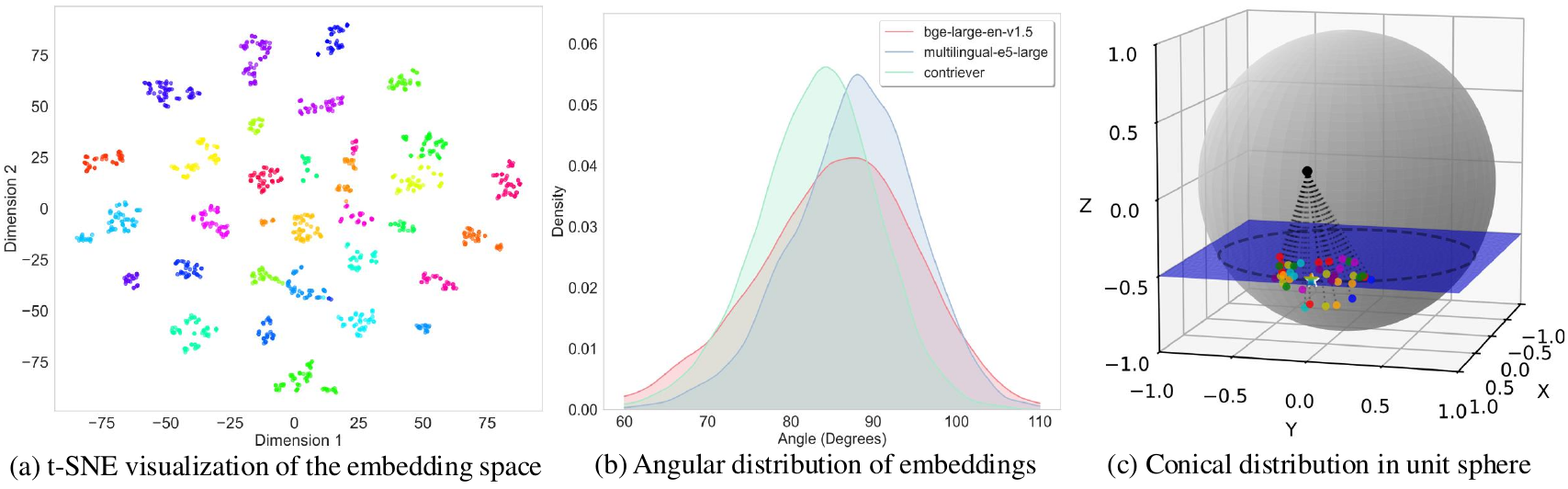}
    \caption{The conical distribution hypothesis.
    (a) t-SNE visualization of queries derived from various documents in the embedding space, illustrating distinct clustering behavior.
    (b) Angular distribution of document and query embeddings, showing the distribution of angles between $ v_d = \mathcal{E}(d) - \mathbb{E}[\mathcal{E}(\mathcal{Q}(d))] $ and $ v_{q_i} = \mathcal{E}(q_i) - \mathbb{E}[\mathcal{E}(\mathcal{Q}(d))] $. 
    The angles form a bell curve just below 90°, supporting that $ v_d $ is approximately orthogonal to each $ v_{q_i}$.
    (c) 3D visualization illustrating the conical distribution of the document (black point) and query (colored points) embeddings  within a unit sphere. 
    The star indicates the queries' cluster center.}
    
    \label{fig_hypo_valid}
    
\end{figure*}
\section{Method}
\subsection{Problem Formulation}
Given a query $q$ and a document corpus $\mathcal{D}=\{d_1, d_2, ..., d_N\}$, our task is to retrieve a subset of $K$ most relevant documents $\mathcal{D}_+=\{d_{i_1},d_{i_2},...,d_{i_K}\}$
through vector search.
We define our embedding model as $\mathcal{E}(\cdot)$, which maps each document $d$ and query $q$ from the textual space $\mathcal{C} $ to a vector space $ \mathbb{R}^r $.
The semantic relevance is quantified by the cosine similarity, defined as: $sim(q, d) = \frac{\mathcal{E}(q)^T\mathcal{E}(d)}{\|\mathcal{E}(q)\| \|\mathcal{E}(d)\|}.$
Furthermore, for each document $d$ in our datasets, we invoke the query generator $\mathcal{Q}(\cdot)$ multiple times to generate $n$ predicted queries $ \{q_i\}_{i=1}^n$, where the cluster center in embedding space is captured by $\mathbb{E}[\mathcal{E}(\mathcal{Q}(d))]$ and Monte Carlo estimation $\overline{\mathcal{E}(\mathcal{Q}(d))}$.

\subsection{QAEncoder}
We first introduce a novel encoding method, $\text{QAE}_{\text{base}}$, which represents the document by the cluster center $\mathbb{E}[\mathcal{E}(\mathcal{Q}(d))]$ of potential queries.
Formally, the representation is defined as follows:
\begin{equation}
\begin{aligned}
\text{QAE}_{\text{base}}(d) & = \mathbb{E}[\mathcal{E}(\mathcal{Q}(d))] 
\\&\approx \overline{\mathcal{E}(\mathcal{Q}(d))} = \frac{1}{n} \sum_{i=1}^{n} \mathcal{E}(q_i).
\label{eq:QAE_base}
\end{aligned}
\end{equation}

To thoroughly illustrate its mechanism, we formally define the \textit{conical distribution hypothesis} and validate its reasonableness with empirical analysis.

\begin{hypothesis}[Conical Distribution Hypothesis]
\label{conical_hypo}
For any document $d$, the potential queries approximately form a single cluster on some hyperplane $\mathcal{H} = \{x \in \mathbb{R}^r \mid w \cdot x = b\}$ in the semantic space, where $ w \in \mathbb{R}^r $ is the normal vector and $ b \in \mathbb{R} $ is the bias term. Furthermore, the document embedding $\mathcal{E}(d)$ lies on the perpendicular line intersecting the cluster center $\mathbb{E}[\mathcal{E}(\mathcal{Q}(d))]$. Formally, the relationship can be expressed as:
$$
\mathcal{E}(d) \approx \mathbb{E}[\mathcal{E}(\mathcal{Q}(d))] + \lambda w,\quad \lambda \in \mathbb{R}.
$$
\end{hypothesis}

We acknowledge that a more realistic model is an oblique cone, not the regular cone. Despite the hypothesis being highly simplified, it clearly captures the dynamics of QAEncoder's alignment. As shown in Fig. \ref{fig_hypo_valid}, the document embedding \(\mathcal{E}(d)\) is an outlier of the cluster of potential queries; whereas \(\text{QAE}_{\text{base}}(d)\) bridges this substantial gap, mathematically denoted by \(\lambda w\). 
We leave more discussions in Appendix \ref{proof_conical_basic} for interested readers.

Nonetheless, the ideal embedding model should not only bring related entries closer but also separate unrelated entries as much as possible. Despite $\text{QAE}_{\text{base}}$ increases document-query similarity, it poses the distinguishability issue.

\subsubsection{Document Fingerprint Strategies}
\begin{figure*}[tb!]
    \centering
    \makebox[\textwidth][c]{\includegraphics[width=\textwidth]{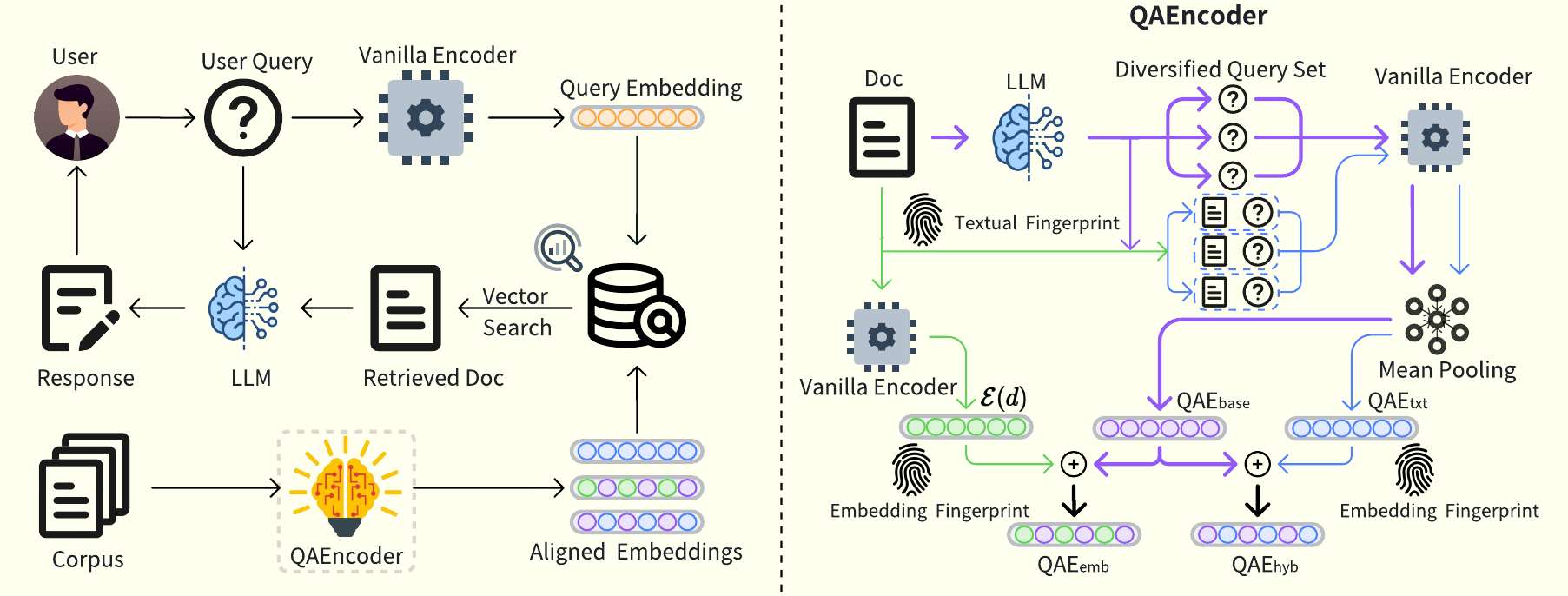}}
    \caption{Architecture of  QAEncoder. \underline{Left:} Corpus documents are embedded using QAEncoder to obtain query-aligned representations for indexing.   User queries are encoded with a vanilla encoder and used to retrieve relevant documents.    
    \underline{Right:} Internal mechanism of QAEncoder.
    QAEncoder  addresses the document-query gap by generating a diverse set of queries for each document to create semantically aligned embeddings.
    Additionally, document fingerprint strategies are employed to ensure document distinguishability.}
    \label{fig:pipeline and architecture}
\end{figure*}

The distinguishability issue arises because incorporating too much query semantics into document representations suppresses their unique characteristics, and renders unrelated documents more similar. To address this issue, we reintroduce the characteristic of documents, intuitively termed \textit{document fingerprints}, which enhance the uniqueness of QAE\textsubscript{base} representations from different perspectives.

\textbf{QAE\textsubscript{base} + Embedding fingerprint =} $ \text{\textbf{QAE}}_{\text{\textbf{emb}}} $.
    The QAE\textsubscript{emb} strategy manipulates within the embedding space and reintroduces unique identity of the original document, i.e. the document embedding $\mathcal{E}(d)$. Specifically, $ \text{QAE}_{\text{emb}}$ considers both the cluster center, $ \text{QAE}_{\text{base}} $, and the document embedding, $ \mathcal{E}(d) $, balancing their contributions using a hyperparameter $\alpha$. The adjusted embedding is formulated as follows:

\begin{equation}
\label{eq_qae_emb}
\begin{aligned}
\text{QAE}_{\text{emb}}(d) &= (1 - \alpha) \cdot \mathcal{E}(d)+ \alpha \cdot \text{QAE}_{\text{base}}(d) \\ &\approx (1 - \alpha) \cdot \mathcal{E}(d)+ \alpha \cdot \frac{1}{n} \sum_{i=1}^{n} \mathcal{E}(q_i).
\end{aligned}
\end{equation}

\textbf{QAE\textsubscript{base} + Textual fingerprint = } $ \text{\textbf{QAE}}_{\text{\textbf{txt}}} $.
    The $ \text{QAE}_{\text{txt}} $ strategy focuses on the textual space and injects the document identity in a more straightforward manner. 
    Let us define the length of text $c$ as $|c|$. Before embedding, each document $d$ is enriched by concatenating it with predicted queries to achieve a length ratio of approximately $\beta$ between the queries and the original document. Then, the final embedding is derived as the average representation of these enriched documents.

\begin{equation}
\label{eq_qae_txt}
\begin{aligned}
      d_i^{*} = \text{concat}&(d, \{q_{j}\}_{j=1}^k),\;
      \text{s.t.} \; | d_i^{*} | \approx (1+\beta) |d|. \\
     & \text{QAE}_{\text{txt}}(d) = \frac{1}{n} \sum\limits_{i=1}^n \mathcal{E}(d_i^{^*}).
\end{aligned}
\end{equation}
Here $k$ is dynamically determined by the constraint $|d_i^{\*}|\approx (1+\beta)|d|$. For each enriched document $d_i^{*}$, we first shuffle predicted queries, and iteratively append them after $d$ until reaching the length limit.

\textbf{Hybrid fingerprint -} $ \text{\textbf{QAE}}_{\text{\textbf{hyb}}} $.
The hybrid approach, QAE\textsubscript{hyb}, seeks to combine the benefits of both QAE\textsubscript{emb} and QAE\textsubscript{txt}.
Although QAE\textsubscript{emb} integrates the document embedding $\mathcal{E}(d)$ and the cluster center $\mathbb{E}[\mathcal{E}(\mathcal{Q}(d))]$ through linear interpolation, inherent differences between these embeddings suggest that a simple linear interpolation should be suboptimal. Therefore, we explore the potential of substituting the document embedding $\mathcal{E}(d)$ in Equation \ref{eq_qae_emb}
with QAE\textsubscript{txt}, which fuses the semantics of both documents and queries.  
\begin{equation}
        \text{QAE}_{\text{hyb}}(d) =  (1 - \alpha) \cdot \text{QAE}_{\text{txt}}(d)+ \alpha \cdot \text{QAE}_{\text{base}}(d).
\end{equation}

In our implementation, all calculated embeddings are normalized for standardized cosine similarity search. Our experiments confirm $\text{QAE}_{\text{emb}}$ and $\text{QAE}_{\text{hyb}}$ outperform $\text{QAE}_{\text{base}}$ and $\text{QAE}_{\text{txt}}$.

\section{Experiments}

\begin{table*}[bt]
\centering
\resizebox{\textwidth}{!}{

 \begin{tabular}{cccccccccc}
 \midrule
    Model & Method & \textbf{AVG} & ArguAna & CQADups. & FEVER & MSMARCO & SciFact & Touche20 & TRECC. \\
    \midrule
    \multicolumn{10}{c}{\textbf{Sparse}} \\
    \midrule
    BM25  & -     & 46.9  & 31.5  & 29.9  & \textbf{75.3} & 22.8  & 66.5  & \textbf{36.7} & 65.6 \\
    DocT5Query & -     & \textbf{49.4} & \textbf{34.9} & \textbf{32.5} & 71.4  & \textbf{33.8} & \textbf{67.5} & 34.7  & \textbf{71.3} \\
    \midrule
    \multicolumn{10}{c}{\textbf{Dense}} \\
    \midrule
    \multirow{2}[0]{*}{dpr} & -     & 26.4  & 17.5  & 15.3  & 56.2  & 17.7  & 31.8  & 13.1  & 33.2 \\
          & QAE\textsubscript{emb}, $\alpha=0.45$ & \textbf{37.1} & \textbf{29.6} & \textbf{22.3} & \textbf{70.9} & \textbf{25.4} & \textbf{40.4} & \textbf{22.9} & \textbf{48.0} \\
          \midrule
    \multirow{2}[0]{*}{contriever} & -     & 38.1  & 37.9  & 28.4  & 68.2  & 20.6  & 64.9  & 19.3  & 27.4 \\
          & QAE\textsubscript{emb}, $\alpha=0.45$ & \textbf{45.8} & \textbf{47.1} & \textbf{33.5} & \textbf{73.1} & \textbf{26.1} & \textbf{70.2} & \textbf{25.2} & \textbf{45.2} \\
          \midrule
    \multirow{2}[0]{*}{contriever-msmarco} & -     & 49.0  & 44.6  & 34.5  & 75.8  & \textbf{40.7} & 67.7  & 20.4  & 59.6 \\
          & QAE\textsubscript{hyb}, $\alpha=0.3,\;\beta=0.75$ & \textbf{54.9} & \textbf{53.9} & \textbf{38.2} & \textbf{82.0} & 39.9  & \textbf{73.6} & \textbf{26.3} & \textbf{70.6} \\
          \midrule
    \multirow{2}[0]{*}{bge-large-en-v1.5} & -     & 58.5  & 63.5  & 42.2  & 87.2  & \textbf{42.5} & 74.6  & 24.8  & 74.8 \\
          & QAE\textsubscript{hyb}, $\alpha=0.15,\;\beta=0.5$ & \textbf{61.8} & \textbf{68.8} & \textbf{45.6} & \textbf{91.5} & 41.2  & \textbf{78.9} & \textbf{28.1} & \textbf{78.2} \\
          \midrule
    \multirow{2}[0]{*}{multilingual-e5-large} & -     & 55.0  & 54.4  & 39.7  & \textbf{82.8} & \textbf{43.7} & 70.4  & 23.1  & 71.2 \\
          & QAE\textsubscript{hyb}, $\alpha=0.15,\;\beta=1.5$ & \textbf{58.0} & \textbf{61.1} & \textbf{44.3} & 82.1  & 43.0  & \textbf{73.9} & \textbf{26.3} & \textbf{75.1} \\
          \midrule
    \multirow{2}[0]{*}{e5-large-v2} & -     & 52.9  & 46.4  & 37.9  & 82.8  & \textbf{43.5} & 72.2  & 20.7  & 66.5 \\
          & QAE\textsubscript{hyb}, $\alpha=0.3,\;\beta=1.0$ & \textbf{57.0} & \textbf{55.1} & \textbf{41.2} & \textbf{86.5} & 42.8  & \textbf{75.3} & \textbf{23.8} & \textbf{74.2} \\
          \midrule
    \multirow{2}[0]{*}{gte-base-en-v1.5} & -     & 59.4  & 63.5  & 39.5  & \textbf{94.8} & \textbf{42.6} & 76.8  & 25.2  & 73.1 \\
          & QAE\textsubscript{hyb}, $\alpha=0.3,\;\beta=0.5$ & \textbf{62.2} & \textbf{68.2} & \textbf{43.7} & 94.2  & 41.9  & \textbf{80.3} & \textbf{29.3} & \textbf{77.5} \\
          \midrule
    \multirow{2}[0]{*}{jina-embeddings-v2-small-en} & -     & 48.9  & 46.7  & 38.0  & 68.0  & 37.3  & 63.9  & 23.5  & 65.2 \\
          & QAE\textsubscript{hyb}, $\alpha=0.15,\;\beta=0.5$ & \textbf{54.2} & \textbf{55.3} & \textbf{42.3} & \textbf{74.3} & \textbf{39.5} & \textbf{67.2} & \textbf{27.2} & \textbf{73.6} \\
          \midrule
    \end{tabular}%

}

\caption{
Retrieval performance on seven BEIR benchmarks (NDCG@10).
Hyperparameters including QAEncoder variants and weight terms $\alpha,\;\beta$ are optimized simultaneously for all datasets in BEIR. See Table \ref{full_classic} for the full table.  
}
\label{tab:classical1}
\end{table*}

\paragraph{Datasets and Metrics.} To rigorously assess the effectiveness of QAEncoder, we employ well-established BEIR benchmark \citep{thakur2021beir}, which contains fifteen publicly available datasets to evaluate the general retrieval performance across diverse domains.
However, classical datasets are frequently utilized for pre-training or fine-tuning embedding models \footnote{Please see the fine-tuning data of \href{https://huggingface.co/datasets/Shitao/bge-m3-data}{bge-m3} and  \href{https://huggingface.co/datasets/hanhainebula/bge-multilingual-gemma2-data}{bge-multilingual-gemma2-9b}.}, involving ArguAna, FEVER, FiQA, HotpotQA, MSMARCO, NQ, SciDocs and so on (all from BEIR). Hence, \textit{classical datasets gradually fall short of objectively reflecting the generalized alignment capabilities for state-of-the-art models}, particularly in a rapidly evolving and updated knowledge base.

Recognizing this limitation, we further test on two latest news datasets, the Chinese dataset CRUD-RAG \citep{CRUD-RAG} and the multilingual dataset FIGNEWS \citep{zaghouani2024fignews} covering English, Arabic, French, Hindi and Hebrew.
For BEIR benchmark, we report NDCG@10 as the evaluation metric like previous works \citep{thakur2021beir,wang2021gpl,meng2022augtriever}. For the latest datasets, we report both MRR@10 and NDCG@10 metrics, capturing both recall and ranking capabilities. We adopt the cheapest GPT-4o-mini as the main query generator, see Appendix \ref{app:pipeline} for query prediction pipeline and cost details.

\paragraph{Hyperparameter Setting.}
For the hyperparameters $\alpha$ and $\beta$, we define the following search spaces: $\alpha \in \{0.0, 0.15, 0.3, 0.45, 0.6, 0.75, 0.9\}$ and $\beta \in \{0.25, 0.5, 0.75, 1.0, 1.25, 1.5\}$. We adopt grid search for QAE\textsubscript{hyb}. See Appendix \ref{hyperparameter_search} for more hyperparameter selection consideration. 

For space reasons, we leave implementation details in Appendix \ref{dataset}, the results on complete BEIR benchmark and FIGNEWS datasets in French, Hindi, and Hebrew in Appendix \ref{full_tables}.

\subsection{Main Results}

We mainly compare QAEncoder against the vanilla encoders, i.e. the backbones. Query-centric methods for sparse retrievers are also presented.  Our comparison involves the following approaches:

\textbf{Sparse retrievers} - BM25 \citep{BM25} and DocT5Query \citep{docT5query}, an improved version of Doc2Query.

\textbf{Dense retrievers} - The state-of-the-art embedding models such as BGE models \citep{BGE} by BAAI, E5 models \citep{wang2023improving} by Microsoft, GTE models \citep{zhang2024mgte} by Alibaba-NLP, Jina models \citep{gunther2023jina} by Jina AI; Other well-known models like Contriever models \citep{Contriever} by Facebook Research, BCEmbedding models \citep{youdao_bcembedding_2023} by NetEase Youdao and the popular Text2Vec models \citep{Text2vec}. For more reference, we also include the seminal dense retriever DPR \citep{karpukhin2020dense}, and models fine-tuned on Quora \citep{thakur2021beir}, a large dataset of question pairs for question-query retrieval.
We integrate them with QAEncoder to bridge the document-query gap.

More details can be found in Appendix \ref{baseline}.
\subsubsection{Performance on Classical Datasets}

As shown in Table \ref{tab:classical1}, 
for sparse retrievers, DocT5Query effectively augments documents with query prediction, and exhibits significant improvements on the BEIR benchmark over the standard BM25.
Dense retrievers gradually outperform sparse retrievers with the scaling-up of model parameter and traning data. For the state-of-the-art embedding models such as BGE, E5, and Jina, integrating them with QAEncoder can lead to robust and generalized alignment, particularly for rare or unseen datasets. For instance, the jina-embeddings-v2-small-en model witnesses a NDCG increase from 65.2 to 73.6 on TRECC dataset; the bge-large-en-v1.5 model's NDCG on SciFact dataset rises from 74.6 to 78.9; the e5-large-v2 model achieves 8.7 NDCG gains on the ArguAna dataset, as the ArguAna dataset is not included in the fine-tuning data \citep{wang2022text}.

Moreover, QAEncoder significantly improves the performance of other well-known models. For example, the contriever model and its fine-tuned version, contriever-msmarco, achieve 17.8 and 11 NDCG gains on TRECC dataset respectively. More data is available in Tab. \ref{full_classic}.

\begin{table*}[!h]
  \centering
\resizebox{\textwidth}{!}{
\begin{tabular}{cccccccc}
\toprule
\multirow{2}[4]{*}{Model} & \multirow{2}[4]{*}{Method} & \multicolumn{2}{c}{FIGNEWS(English)} & \multicolumn{2}{c}{FIGNEWS(Arabic)} & \multicolumn{2}{c}{CRUD-RAG(Chinese)} \\
\cmidrule{3-8}      &       & \multicolumn{1}{l}{MRR@10} & \multicolumn{1}{l}{NDCG@10} & \multicolumn{1}{l}{MRR@10} & \multicolumn{1}{l}{NDCG@10} & \multicolumn{1}{l}{MRR@10} & \multicolumn{1}{l}{NDCG@10} \\
\midrule
\multirow{2}[2]{*}{bge-m3} & -     & 74.4  & 78.7  & 77.8  & 80.9  & 47.5  & 48.6 \\
      & QAE\textsubscript{txt}, $\beta=1.5$ & \textbf{77.2} & \textbf{81} & \textbf{80.2} & \textbf{83.1} & \textbf{51.4} & \textbf{52.5} \\
\midrule
\multirow{2}[2]{*}{multilingual-e5-small} & -     & 71    & 75.1  & 74.1  & 77.4  & 44.6  & 46.0 \\
      & QAE\textsubscript{hyb}, $\alpha=0.3,\;\beta=0.5$ & \textbf{74.6} & \textbf{78.5} & \textbf{78.9} & \textbf{81.6} & \textbf{50.6} & \textbf{51.6} \\
\midrule
\multirow{2}[2]{*}{multilingual-e5-base} & -     & 74.8  & 78.1  & 72.3  & 76    & 47.0  & 48.2 \\
      & QAE\textsubscript{emb}, $\alpha=0.3$ & \textbf{77.6} & \textbf{81.3} & \textbf{77.2} & \textbf{80.3} & \textbf{51.2} & \textbf{52.3} \\
\midrule
\multirow{2}[2]{*}{multilingual-e5-large} & -     & 73.9  & 77.8  & 76.7  & 80.2  & 46.9  & 48.3 \\
      & QAE\textsubscript{hyb}, $\alpha=0.15,\;\beta=1.25$ & \textbf{77.1} & \textbf{80.6} & \textbf{82.2} & \textbf{85.1} & \textbf{51.5} & \textbf{52.7} \\
\midrule
\multirow{2}[2]{*}{gte-multilingual-base} & -     & 65.5  & 70.4  & 73.4  & 76.8  & 45.3  & 46.8 \\
      & QAE\textsubscript{hyb}, $\alpha=0.15,\;\beta=1.5$ & \textbf{75.5} & \textbf{79.5} & \textbf{76.2} & \textbf{79.1} & \textbf{49.4} & \textbf{51.0} \\
\midrule
\multirow{2}[2]{*}{mcontriever} & -     & 32.9  & 36.7  & 40.3  & 44.7  & 39.2  & 41.6 \\
      & QAE\textsubscript{hyb}, $\alpha=0.45,\;\beta=1.25$ & \textbf{61.4} & \textbf{65.9} & \textbf{68.3} & \textbf{72.1} & \textbf{51.3} & \textbf{52.4} \\
\midrule
\multirow{2}[2]{*}{bce-embedding-base-v1} & -     & 59.1  & 63.8  & -     & -     & 42.0  & 44.0 \\
      & QAE\textsubscript{hyb}, $\alpha=0.3,\;\beta=0.5$ & \textbf{66.8} & \textbf{71.1} & -     & -     & \textbf{49.7} & \textbf{51.0} \\
\midrule
\multirow{2}[2]{*}{text2vec-base-multilingual} & -     & 38.7  & 43.6  & 27.8  & 31.9  & 9.7   & 10.6 \\
      & QAE\textsubscript{emb}, $\alpha=0.75$ & \textbf{55.4} & \textbf{59.9} & \textbf{51.5} & \textbf{55.4} & \textbf{32.1} & \textbf{34.1} \\
\midrule
\multirow{2}[2]{*}{quora-distilbert-multilingual} & -     & 39.8  & 44.2  & 28.2  & 32.1  & 21.8   & 23.5 \\
      & QAE\textsubscript{emb}, $\alpha=0.6$ & \textbf{50.5} & \textbf{54.7} & \textbf{38.7} & \textbf{42.8} & \textbf{34.4} & \textbf{36.8} \\
\bottomrule
\end{tabular}%

}
\caption{
Retrieval performance on the latest datasets FIGNEWS and CRUD-RAG.
Hyperparameters including QAEncoder variants and weight terms $\alpha,\;\beta$ are optimized simultaneously for all lastest datasets. 
We leave the full table in Tab. \ref{full_latest}, and results on monolingual and bilingual embedding models in Tab. \ref{cn_en} for interested readers.
}
  \label{tab:latest1}%
\end{table*}%

\subsubsection{Performance on Latest Datasets}
In scenarios such as search engine, financial analysis, and news QA, large volumes of new data constantly emerge and are indexed into retrieval base for accurate and up-to-date response. Hence, the alignment capability for previously unseen user queries and relevant documents is crucial for embedding models in RAG systems. We experiment on the latest news datasets, FIGNEWS and CRUD-RAG, to avoid data leakage and mimic the real-world scenarios.  
As illustrated in Table \ref{tab:latest1}, for latest datasets, QAEncoder significantly improves across both state-of-the-art embedding models and other well-known models. E.g., the gte-multilingual-base model and the mcontriever model's MRR metrics increase from 65.5 to 75.5, and 32.9 to 61.4 respectively on FIGNEWS(English) dataset. The multilingual-e5-large model's MRR increase from 76.7 to 82.2 on FIGNEWS(Arabic).
\textcolor{black}{
Additionally, the text2vec-base-multilingual model's MRR on CRUD-RAG dataset rises from 9.7 to 32.1, while the bge-m3 model's MRR improves from 47.5 to 51.4.}
These results remarkably confirm QAEncoder's generalized alignment capability across various embedding models and languages. 
See Tables \ref{full_latest} and \ref{cn_en} for more embedding models and datasets.

\subsection{Analysis and Discussion}
For more comprehensive assessments, we analyze various QAEncoder ablations, i.e. QAE\textsubscript{emb}, QAE\textsubscript{txt}, QAE\textsubscript{hyb} and QAE\textsubscript{naive}, which directly stores predicted queries. We also evaluate QAEncoder's robustness with respect to the query generator. Finally, we discuss the relationship between QAEncoder and both training-based and document-centric methods.

\subsubsection{Ablations of QAEncoder}
\begin{figure*}[tb!]
    \centering
    \includegraphics[width=0.325\linewidth]{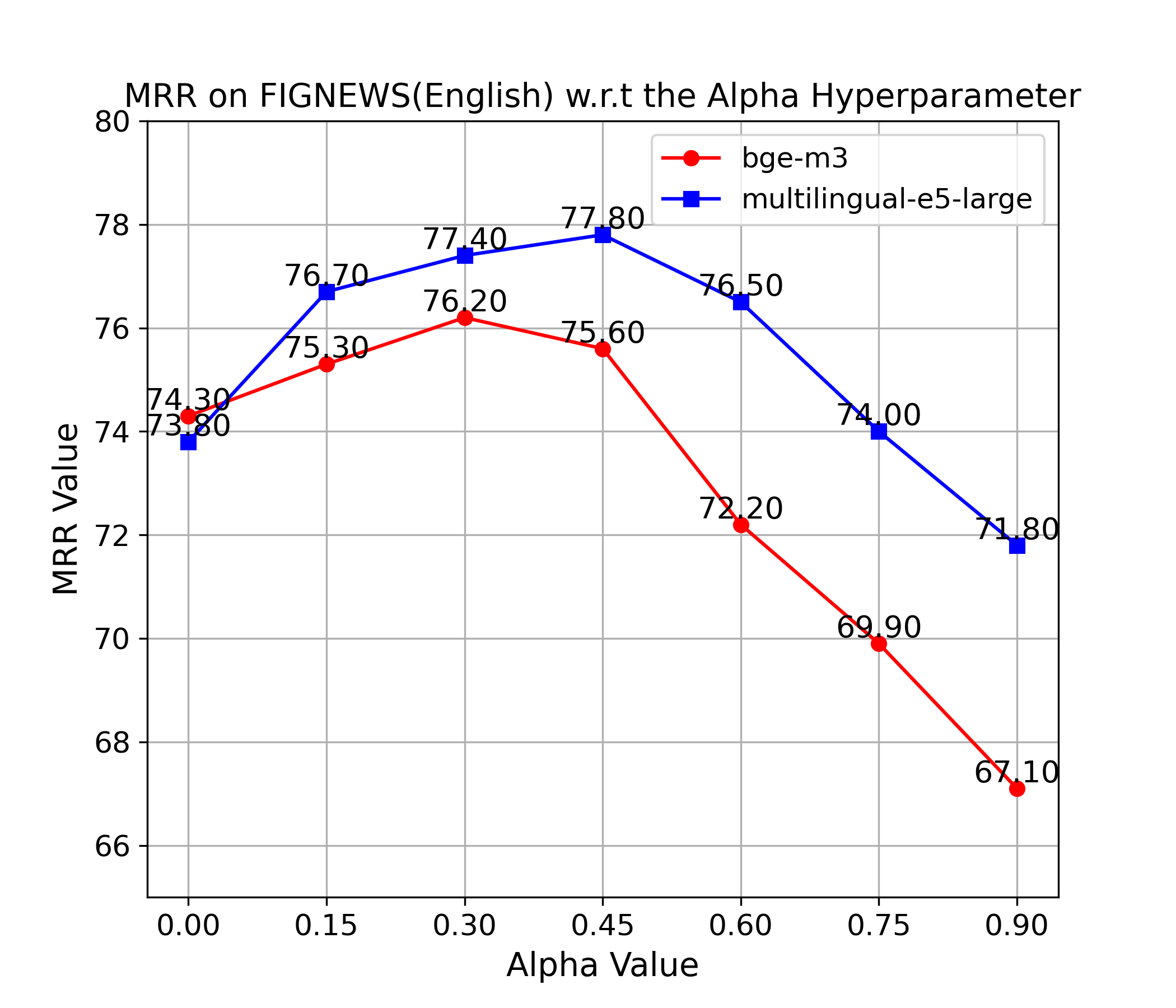}
    \includegraphics[width=0.325\linewidth]{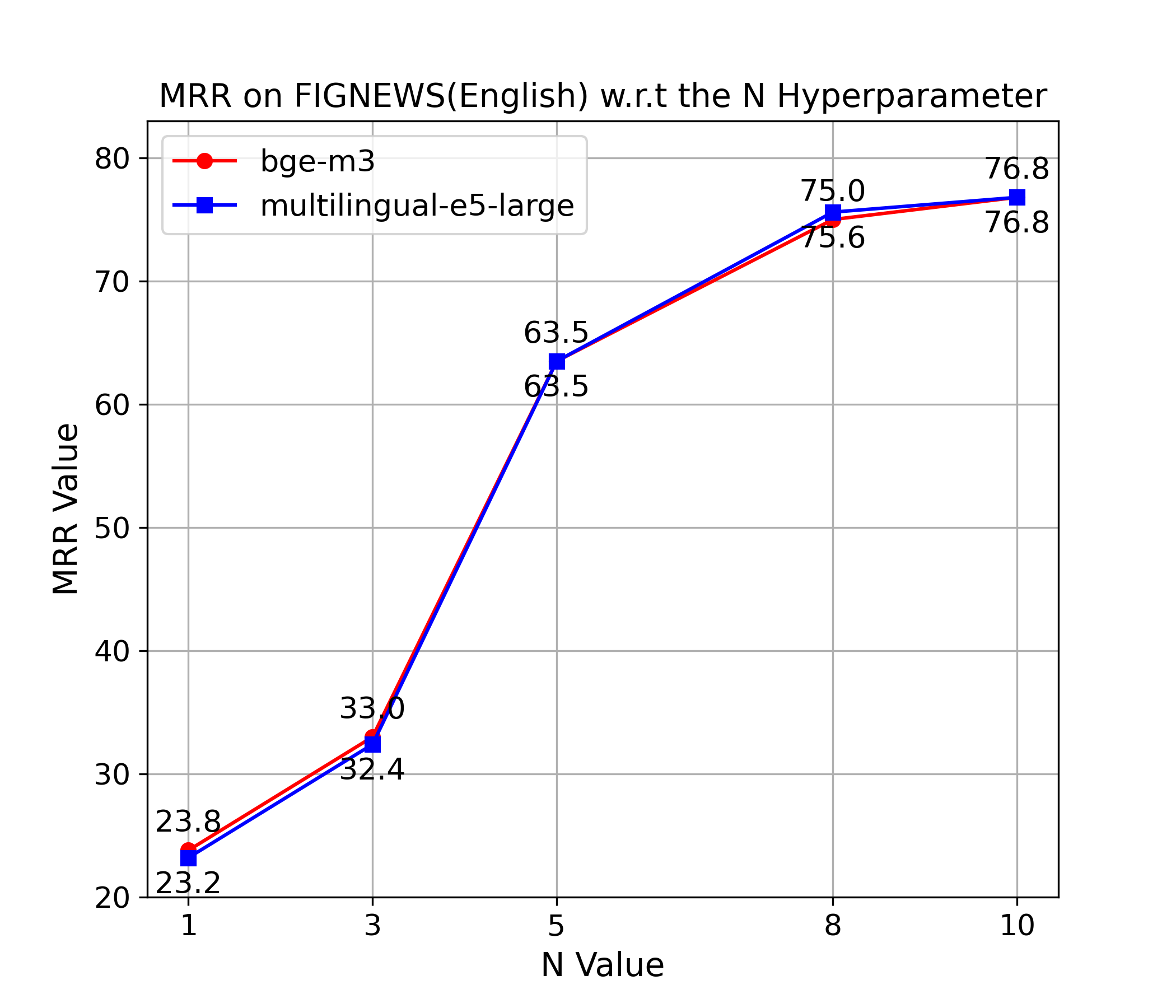}
    \includegraphics[width=0.325\linewidth]{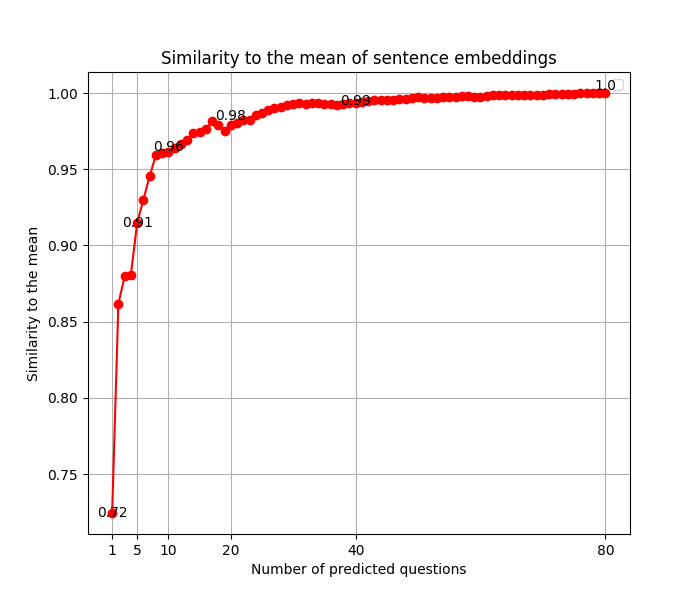}
    \caption{
    \underline{Left}: The impact of varying $\alpha$ values for QAE\textsubscript{emb}.
    \underline{Middle}: The effect of varying the number of predicted queries for QAE\textsubscript{naive}, with MRR improving as $n$ increases, approaching stability at $n = 10$. The curves for different models are mostly identical as the matching are largely driven by predicted queries. \underline{Right}: The convergence speed for Monte Carlo estimation of QAE\textsubscript{base}. Please refer to Appendix \ref{append_cost} and Fig. \ref{fig:monte_carol} for more cost details.
    }
    \label{fig:Ablation_alpha_naive}
\end{figure*}

\begin{table*}[tb!]
  \centering

    \resizebox{\textwidth}{!}{
    
\begin{tabular}{cccccccc}
\toprule
\multirow{2}[4]{*}{Model} & \multirow{2}[4]{*}{Method} & \multicolumn{2}{c}{FIGNEWS(English)} & \multicolumn{2}{c}{FIGNEWS(Arabic)} & \multicolumn{2}{c}{CRUD-RAG(Chinese)} \\
\cmidrule{3-8}      &       & \multicolumn{1}{l}{MRR@10} & \multicolumn{1}{l}{NDCG@10} & \multicolumn{1}{l}{MRR@10} & \multicolumn{1}{l}{NDCG@10} & \multicolumn{1}{l}{MRR@10} & \multicolumn{1}{l}{NDCG@10} \\
\midrule
\multirow{4}[2]{*}{bge-m3} & QAE\textsubscript{emb}, $\alpha=0.3$ & 76.4  & 80.5  & 80.1  & 82.9  & 51.3  & 52.4 \\
      & QAE\textsubscript{txt}, $\beta=1.5$ & 77.2  & 81    & 80.2  & 83.1  & 51.4  & 52.5 \\
      & QAE\textsubscript{hyb}, $\alpha=0.15,\;\beta=1.5$ & \textbf{77.4} & \textbf{81.1} & \textbf{80.6} & \textbf{83.4} & \textbf{51.7} & \textbf{52.7} \\
      & QAE\textsubscript{naive}, n=10 & 76.9  & 79.9  & 77.1  & 79.7  & 47.0  & 48.1 \\
\midrule
\multirow{4}[2]{*}{multilingual-e5-large} & QAE\textsubscript{emb}, $\alpha=0.45$ & \textbf{77.9} & \textbf{81.4} & 79.8  & 83    & \textbf{51.9} & \textbf{52.9} \\
      & QAE\textsubscript{txt}, $\beta=1.5$ & 75.6  & 79.2  & 80.9  & 84.1  & 51.0  & 52.3 \\
      & QAE\textsubscript{hyb}, $\alpha=0.15,\;\beta=1.25$ & 77.1  & 80.6  & \textbf{82.2} & \textbf{85.1} & 51.5  & 52.7 \\
      & QAE\textsubscript{naive}, n=10 & 77.5  & 80.3  & 76.5  & 79.4  & 46.5  & 47.7 \\
\bottomrule
\end{tabular}%

    }

  \caption{Performance comparison of QAEncoder variants on the latest datasets.
  Hyperparameters are optimized simultaneously for all the latest datasets. $n$ indicates the number of predicted queries in QAE\textsubscript{naive}. 
  }
  \label{tab:ablation1}
\end{table*}
We present the performance comparison of QAEncoder variants on two state-of-the-art embedding models in Table \ref{tab:ablation1}.
Generally, QAE\textsubscript{hyb} and QAE\textsubscript{emb} outperform the QAE\textsubscript{txt} and QAE\textsubscript{naive} approaches. 
For instance, for the bge-m3 model, QAE\textsubscript{hyb} consistently outperforms other variants. Conversely, the multilingual-e5-large model performs best with QAE\textsubscript{emb}. However, the best performance differences between QAE\textsubscript{emb}, QAE\textsubscript{txt}, and QAE\textsubscript{hyb} are not substantial, demonstrating the robustness of our approach to hyperparameter variations.
Regarding QAE\textsubscript{naive}, it evidently underperforms other ablations, despite storing 10 times the number of embedding vectors. This leads to unacceptable storage management overhead and recall latency in large-scale production systems.
We provide more granular ablation experiments in Fig. \ref{fig:Ablation_alpha_naive}, as well as the convergence speed of QAE\textsubscript{base}'s Monte Carlo estimation. See Table \ref{full:ablation} for the full table.

\subsubsection{Robustness w.r.t. Query Generator}
\begin{figure}[H]
    \centering
    \includegraphics[width=\linewidth]{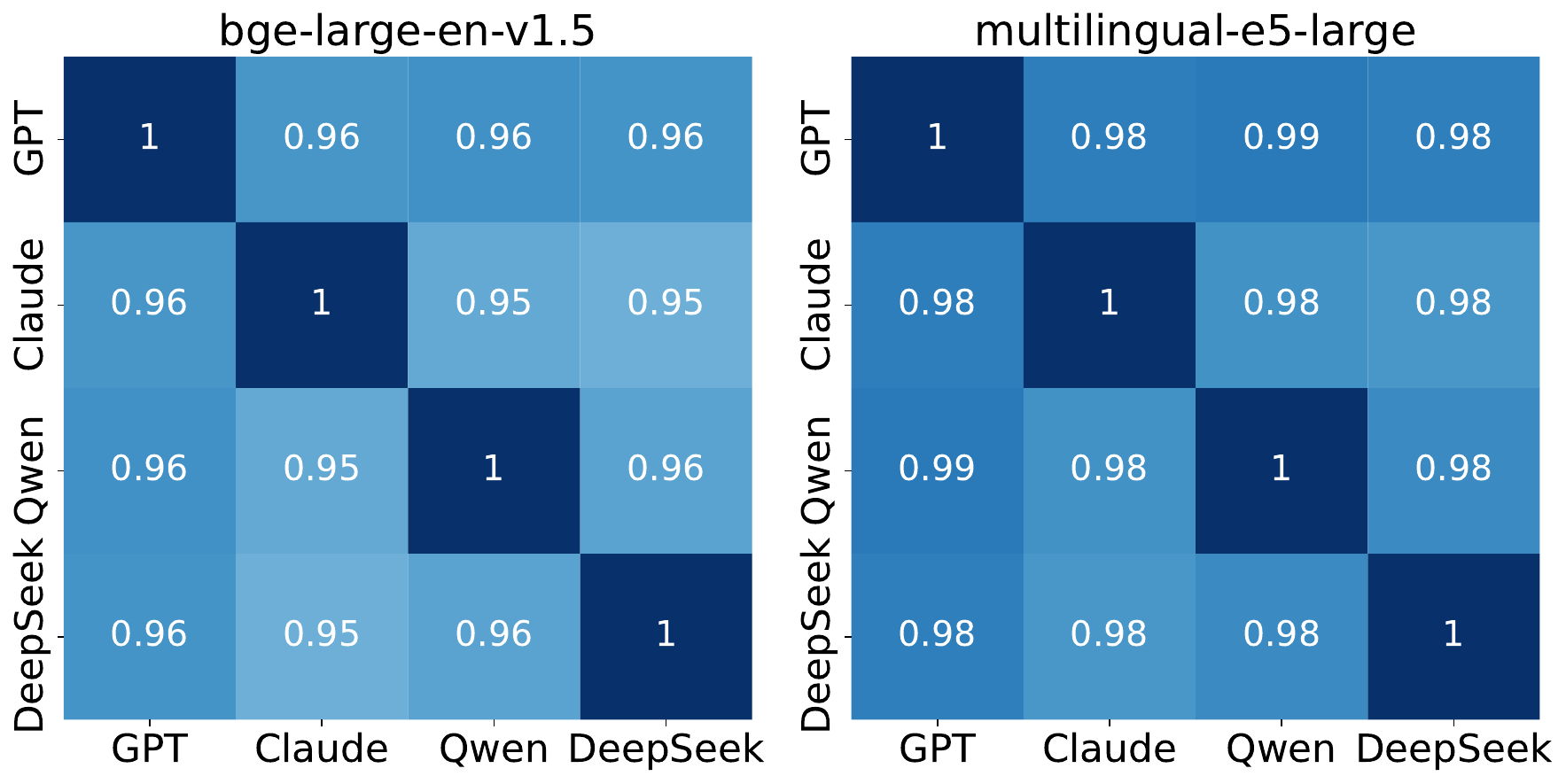}
    \caption{Similarity matrices of QAE$_{\text{base}}$ with different query generators (average value on 100 documents and 10 queries per document), exhibiting high consistency.}
    \label{fig:robustness_bge_e5}
\end{figure}

The calculation of QAE$_{\text{base}} \approx \frac{1}{n} \sum_{i=1}^{n} \mathcal{E}(q_i)$ is largely robust to query generators, thanks to the nature of Monte Carlo estimation \citep{hsu1947complete}. As shown in Fig. \ref{fig:robustness_bge_e5}, for any given document and embedding model (e.g. BGE and E5), the QAE$_{\text{base}}$ representations derived from different query generators are quite consistent. For example, the worst cosine similarity remains as high as 0.95 for the bge-large-en-v1.5 encoder. Note that QAE\textsubscript{emb}, QAE\textsubscript{txt}, and QAE\textsubscript{hyb} integrate document fingerprints into QAE\textsubscript{base}, and thus have better query generator independence.

As shown in Tab. \ref{mrr_consistent}, we adopted GPT-4o-mini to generate test queries and various LLMs as generators for query prediction.
For QAE\textsubscript{emb}, GPT-4o-mini shows a slight advantage over other predictors. But for QAE\textsubscript{txt} and QAE\textsubscript{hyb}, the performance is largely consistent, highlighting not only the importance of document fingerprints but also the robustness of QAEncoder. Interestingly, Claude beats GPT on QAE\textsubscript{txt} with MRR values 73.5 and 72.7 respectively, while the vanilla BGE is 66.1. The full names of abbreviations are in left Appendix \ref{app:pipeline}.
\begin{table}[H]
    \centering
    \resizebox{\columnwidth}{!}{
    \begin{tabular}{cccccc}
        \toprule[1pt]
        Model & Method  & GPT & Claude & Qwen & DeepSeek \\ \midrule
        \multirow{3}[2]{*}{bge-large-en-v1.5}
        & QAE\textsubscript{emb}, $\alpha=0.5$ & 72.2 & 71.6 & 71.0 & 70.5 \\ 
        & QAE\textsubscript{txt}, $\beta=1.5$ & 72.7 & 73.5 & 72.6 & 72.5 \\ 
        & QAE\textsubscript{hyb}, $\alpha=0.15, \beta=1.5$ & 73.7 & 73.5 & 73.2 & 73.3 \\ 
        \midrule
        \multirow{3}[2]{*}{multilingual-e5-large} 
        & QAE\textsubscript{emb}, $\alpha=0.5$ & 77.6 & 76.1 & 75.8 & 75.9 \\ 
        & QAE\textsubscript{txt}, $\beta=1.5$ & 75.1 & 74.9 & 74.7 & 74.6 \\ 
        & QAE\textsubscript{hyb}, $\alpha=0.15, \beta=1.5$ & 77.0 & 76.5 & 76.6 & 76.6 \\ 
        \bottomrule[1pt]
    \end{tabular}
    }
    \caption{MRR@10 on FIGNEWS(English) with GPT-generated test queries and different query predictors.
    The performance is largely robust to query predictors.
    \label{mrr_consistent}}
\end{table}

\begin{table*}[tb!]

  \centering

    \resizebox{\textwidth}{!}{

\begin{tabular}{cccccccc}
\toprule
\multirow{2}[4]{*}{Model} & \multirow{2}[4]{*}{Method} & \multicolumn{2}{c}{FIGNEWS(English)} & \multicolumn{2}{c}{FIGNEWS(Arabic)} & \multicolumn{2}{c}{CRUD-RAG(Chinese)} \\
\cmidrule{3-8}      &       & \multicolumn{1}{l}{MRR@10} & \multicolumn{1}{l}{NDCG@10} & \multicolumn{1}{l}{MRR@10} & \multicolumn{1}{l}{NDCG@10} & \multicolumn{1}{l}{MRR@10} & \multicolumn{1}{l}{NDCG@10} \\
\midrule
\multirow{8}[2]{*}{mcontriever} & -     & 32.9  & 36.7  & 40.3  & 44.7  & 39.2  & 41.6 \\
      & QAE\textsubscript{hyb}, $\alpha=0.45,\;\beta=1.25$ & 61.4  & 65.9  & 68.3  & 72.1  & 51.3  & 52.4 \\
      & GPL$^{\dagger}$   & 68.3  & 73.6  & 72.1  & 75.18 & \textbf{52.8} & \textbf{55.9} \\
      & MS$^{\dagger}$    & 66.1  & 70.6  & 70.2  & 73.7  & 46.5  & 47.8 \\
      & MS$^{\dagger}$ + QAE\textsubscript{hyb},  $\alpha=0.3,\;\beta=0.75$ & \textbf{72.3} & \textbf{76.8} & \textbf{77.3} & \textbf{80.5} & 51.2  & 52.4 \\
      & QA-RAG$^{\ddagger}$  & 31.1  & 34.4  & 42.3  & 46.8  & 43.8  & 45.8 \\
      & Query2Doc$^{\ddagger}$ & 25.1  & 29.0  & 34.5  & 38.9  & 35.7  & 37.2 \\
      & HyDE$^{\ddagger}$  & 25    & 27.9  & 35.7  & 41.9  & 36.7  & 38.7 \\
\midrule
\multirow{8}[2]{*}{multilingual-e5-large} & -     & 73.9  & 77.8  & 76.7  & 80.2  & 46.9  & 48.3 \\
      & QAE\textsubscript{hyb}, $\alpha=0.15,\;\beta=1.25$ & \textbf{77.1} & \textbf{80.6} & \textbf{82.2} & \textbf{85.1} & 51.5  & 52.7 \\
      & GPL$^{\dagger}$    & 75.2  & 78.9  & 79.4  & 82.3  & \textbf{53.6} & \textbf{56.3} \\
      & INS$^{\dagger}$    & 67    & 71.4  & 75    & 78.2  & 43.7  & 45.2 \\
      & INS$^{\dagger}$  + QAE\textsubscript{hyb},  $\alpha=0.15,\;\beta=1.5$ & 75.6  & 79.8  & 80.8  & 83.7  & 51.4  & 52.4 \\
      & QA-RAG$^{\ddagger}$ & 73.3  & 76.5  & 72.8  & 75.6  & 45.8  & 46.5 \\
      & Query2Doc$^{\ddagger}$ & 63.4  & 68.2  & 66.3  & 72.8  & 42.0  & 43.1 \\
      & HyDE$^{\ddagger}$  & 63.6  & 68.3  & 68.3  & 74.1  & 42.3  & 43.6 \\
\bottomrule
\end{tabular}%

    }
  
  \caption{Performance comparison of QAEncoder with training-based and document-centric methods on the latest datasets FIGNEWS and CRUD-RAG. Hyperparameters are optimized simultaneously for all latest datasets.
  $\dagger$ indicates the training-based methods. MS$^{\dagger}$ represents fine-tuning on MSMARCO, i.e. the mcontriever-msmarco model; INS$^{\dagger}$ represents instruction-tuning, i.e. the multilingual-e5-large-instruct model \citep{wang2023improving}. $\ddagger$ indicates the document-centric methods. See Table \ref{full:dis} for the full table.
  }
  \label{tab:discuss1}
\end{table*}

\subsubsection{Training-based and Document-centric Methods}
Training-based approaches mainly include two types: fine-tuning on QA datasets (domain adaptation) and fine-tuning on multi-task instruction datasets. We select mcontriever-msmarco and multilingual-e5-large-instruct as representative methods respectively. 
To unveil the catastrophic forgetting issue of training-based methods, we incorporate GPL \cite{wang2021gpl}, which predicts queries, mines hard negative samples, and distills the re-ranker for unsupervised domain adaptation.

For multi-domain adaptation, we fine-tune on FIGNEWS and CRUD-RAG datasets iteratively to simulate knowledge update. While GPL bridges the document-query gap with notable improvements on CRUD-RAG dataset, the catastrophic forgetting issue is serious: GPL improves marginally on FIGNEWS datasets, while QAEncoder works robustly due to the training-free nature.

Training-based and query-centric methods operate at training time and indexing time, respectively. Therefore, integrating these approaches could lead to more improvements. As illustrated in Table \ref{tab:discuss1}, both types of fine-tuned models significantly benefit from the QAEncoder. For instance, the mcontriever-msmarco model improves MRR from 70.2 to 77.3 on FIGNEWS(Arabic); the multilingual-e5-large-instruct model's MRR increases 8.6 and 7.7 MRR points on the FIGNEWS(English) and CRUD-RAG(Chinese) datasets, respectively. 

For document-centric methods, we consider HyDE \citep{HyDE}, Query2Doc \citep{Query2doc}, and QA-RAG \citep{QA-RAG} for comparison. Note that the pseudo-document generator in QA-RAG is fine-tuned on medical QA datasets, here we adopt out-of-box GPT-4o-mini. 
The widely-reported hallucination phenomenon on the latest datasets is confirmed \citep{Query2doc}. 
As shown in Tab. \ref{tab:discuss1}, the retrieval performance of HyDE and Query2Doc heavily decreases for all the latest datasets, attributed to the hallucination of pseudo-document generation. Although QA-RAG alleviates hallucination by its two-way retrieval and re-ranking mechanism, it still underperforms. Besides, the LLM invocation for pseudo-documents is both costly and time-consuming. In our case, the time for single LLM invocation is more than 2000ms while the time for vector search is less than 10ms. These highlight the irreplaceable importance and practicality of QAEncoder method.

\section{Conclusion}
In this paper, we propose QAEncoder to bridge the document-query gap from the query-centric perspective---a novel, training-free and pioneering approach. QAEncoder replaces document embeddings with the expectation of query embeddings, theoretically supported by the conical distribution hypothesis and practically enhanced by document fingerprint strategies.
Extensive experiments are conducted on both classical BEIR benchmark suite and the latest news datasets, covering 20+ embedding models and 6 languages. QAEncoder demonstrates robust generalization, diverse query handling, and compatibility with existing RAG architectures and training-based methods.

\section*{Acknowledgments}
This work is supported by the National Key R\&D Program of China (2024YFA1014003), National Natural Science Foundation of China (92470121, 62402016), CAAI-Ant Group Research Fund, and High-performance Computing Platform of Peking University.

\newpage

\section*{Limitations}
Despite the benefits of QAEncoder, there are still ongoing works for further improvements:
\begin{itemize}
    \item The current mean pooling could be overly simplistic and limits the performance improvement. The multi-cluster version such as Gaussian mixture models and multi-vector representation could be explored.
    \item Since the query generator mainly generates simple queries, out-of-domain issues with complex, multi-hop queries could happen. However, query optimization has become increasingly prevalent to refine the original user query, making it more suitable for retrieval module. Practically, query decomposition and rewriting can be integrated to break down complex queries into simpler sub-queries and to rewrite these sub-queries into the in-domain style, respectively. Therefore, we focus QAEncoder on the core retrieval module, and leave the handling of complex queries to auxiliary strategies.
    \item The 5W1H framework captures our core design principles and serves as the motivation for our work, primarily applicable to narrative or factual texts. Without loss of generality, QAEncoder estimates the cluster center of potential queries via Monte Carlo method, where the type of query is arbitrary and not constrained to 5W1H. We acknowledge there could be the worst case, user query may deviate from predicted ones, i.e. the outlier. Therefore, more fallback mechanisms beyond document fingerprints possess research value.
    \item There is also a risk of data leakage for property information when predicting queries via API calls. The query prediction process may incur underlying security vulnerabilities.
\end{itemize}
We aim to keep QAEncoder simple-yet-effective, and leave these problems in the future research.

\bibliography{acl_latex}

\clearpage
\appendix
\section*{Appendix}
\label{sec:appendix}
\startcontents[sections]
\printcontents[sections]{l}{1}{\setcounter{tocdepth}{3}}

\section{Conical Distribution Hypothesis}
\label{conial_dis_appendix}

\subsection{Case Studies}
\begin{figure*}
    \centering
    \includegraphics[width=0.8\linewidth]{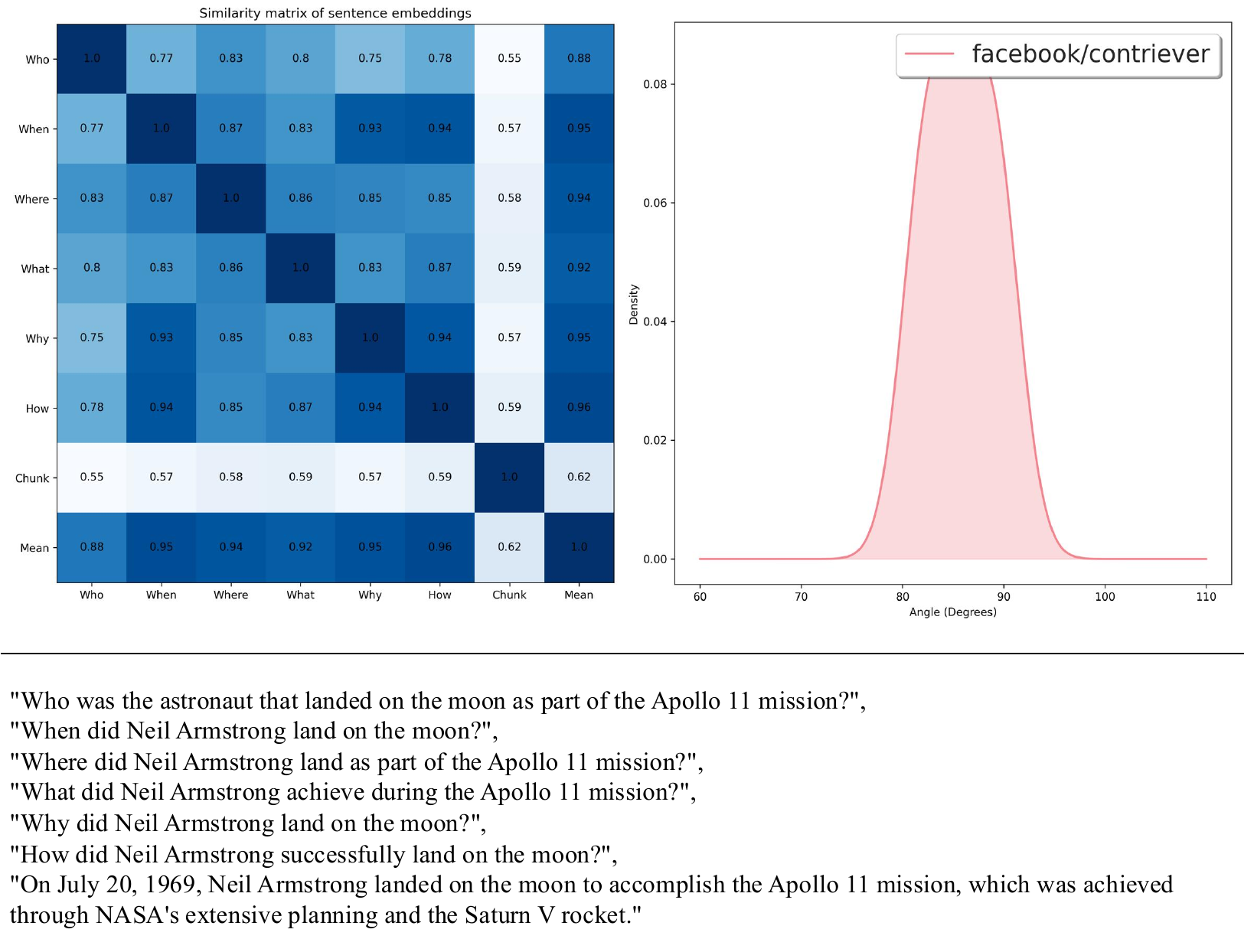}
    \caption{Similarity matrix and angle distribution on \textbf{Armstrong example with contriever encoder}.}
    \label{fig:hypo_case_studies1}
\end{figure*}

\begin{figure*}
    \centering
    \includegraphics[width=0.8\linewidth]{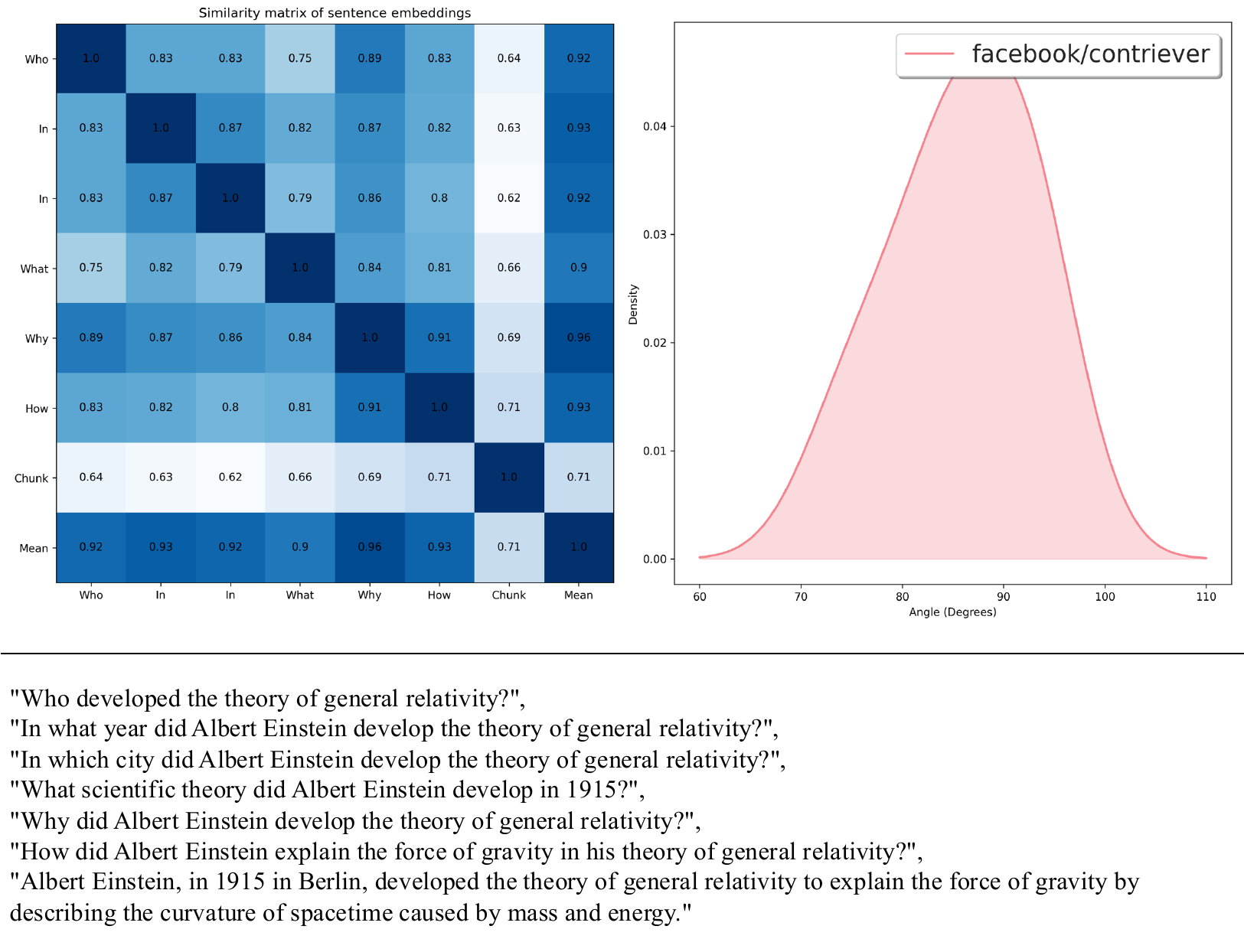}
    \caption{Similarity matrix and angle distribution on \textbf{Einstein example with contriever encoder}.}
    \label{fig:hypo_case_studies2}
\end{figure*}

\begin{figure*}
    \centering
    \includegraphics[width=0.8\linewidth]{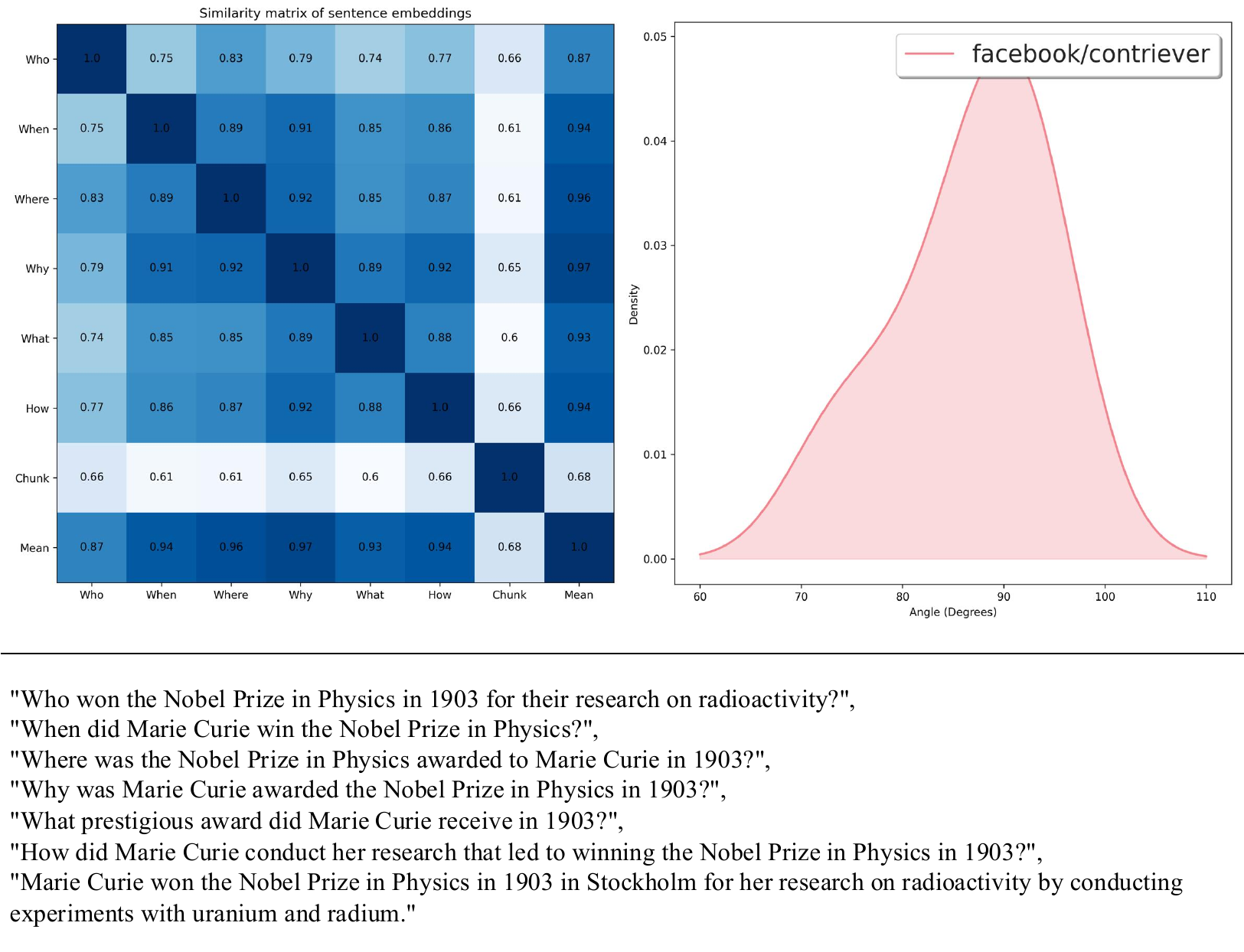}
    \caption{Similarity matrix and angle distribution on \textbf{Curie example with contriever encoder}.}
    \label{fig:hypo_case_studies3}
\end{figure*}

\begin{figure*}
    \centering
    \includegraphics[width=0.8\linewidth]{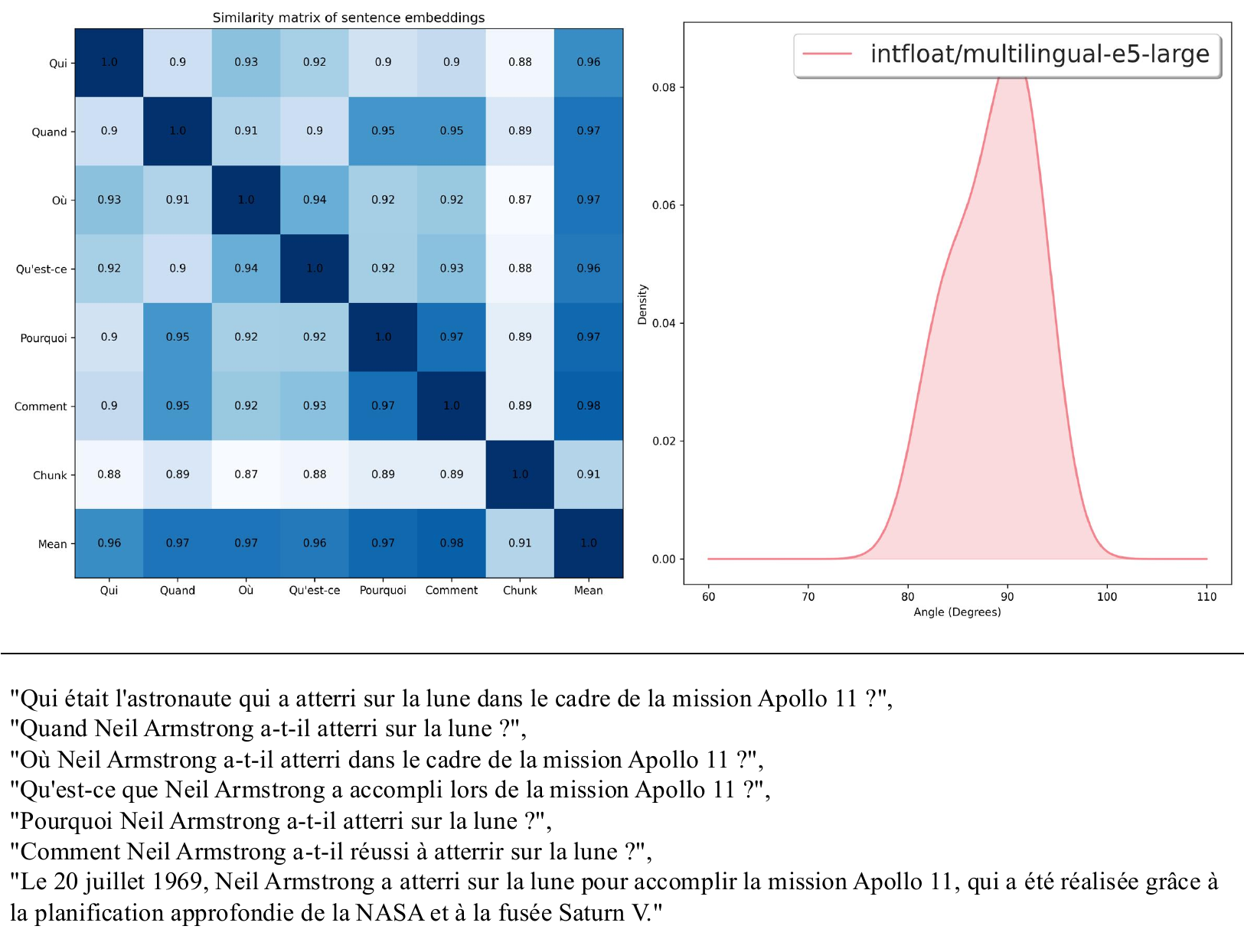}
    \caption{Similarity matrix and angle distribution on \textbf{Armstrong example in French with multilingual-e5-large}.}
    \label{fig:hypo_case_studies4}
\end{figure*}

Here we provide several concrete case studies in Fig. \ref{fig:hypo_case_studies1}, \ref{fig:hypo_case_studies2}, \ref{fig:hypo_case_studies3}, \ref{fig:hypo_case_studies4}. 
The conclusions are highly consistent with experiments in the main text, across embedding models and languages. Please refer to our repository for related scripts and reproduction. 
\subsection{Proof of Conical Distribution Hypothesis}
\label{proof_conical_basic}
This subsection provides the proof of the Conical Distribution Hypothesis, which proposes that potential queries form a distinct cluster on a hyperplane in semantic space.

\begin{proof}
Our validation is structured from three core aspects:
\begin{itemize}
    \item 
\textbf{Single-cluster sub-hypothesis verification.}
 As illustrated in Fig. \ref{fig_hypo_valid}(a), we validate the single-cluster sub-hypothesis by visualizing the embedding space using t-SNE dimensionality reduction techniques. This visualization displays that the predicted queries for each document form distinct and cohesive clusters (different colored). And these clusters are notably distant from the clusters of other documents, thereby supporting the single-cluster sub-hypothesis.
    \item 
\textbf{Perpendicular sub-hypothesis verification.}
To further assess the perpendicular sub-hypothesis, let $v_d =  \mathcal{E}(d)- \mathbb{E}[\mathcal{E}(\mathcal{Q}(d))]$  and $v_{q_i} =  \mathcal{E}(q_i) - \mathbb{E}[\mathcal{E}(\mathcal{Q}(d))]$ be the vectors from the cluster center to the document embedding and the individual query embedding, respectively. As illustrated in Fig. \ref{fig_hypo_valid}(b), the degree distribution between vector $v_d$ and vector $v_{q_i}$ exhibits a bell-shaped curve. The mean value is slightly less than 90 degrees, and the primary range of distribution lies between 75 and 100 degrees, which confirms that $v_d$ is approximately orthogonal to each $v_{q_i}$ and can be regarded as the normal vector to some hyperplane $\mathcal{H}$. 
    \item
\textbf{Conical distribution in unit sphere demonstration.}
Finally, we illustrate the highly simplified conical distribution hypothesis within the unit sphere embedding space, as most embedding models utilize normalized embedding vectors. As depicted in Fig. \ref{fig_hypo_valid}(c), the embeddings of potential queries form a cluster on the surface of the unit sphere, with each point color-coded. The center of the cluster is indicated by a star, while the document embedding is represented by a black point positioned above the cluster. It is evident that these elements form a distorted cone, aligning with the above hypothesis and the degree distribution experiment. 
\end{itemize}
\end{proof}

Furthermore, when the stronger hypothesis assuming the cluster follows the Gaussian distribution is adopted, more quantitative analysis results can be derived.

\subsection{Strong Conical Distribution Hypothesis}
\label{Strong Conical Distribution Hypothesis}
In this subsection, we further substantiate the original hypothesis that the potential queries adhere to a Gaussian distribution:
For any document $d$, the potential queries in the embedding space approximately follow a Gaussian distribution, characterized by a mean $\mu$ and covariance matrix $\Sigma$. Refer to Appendix \ref{Normality_Test} for detailed validation.

\subsubsection{Main theorem}
Building on this Gaussian assumption, we derive bounds on the cosine similarity between potential query embeddings and both the document embedding and the mean vector. 
\begin{theorem}(Concentration Inequalities for Cosine Similarities in  Embedding Spaces) Let $\mathbf{q} \sim \mathcal{N}(\mu, \Sigma)$ denote a random vector representing the distribution of potential queries of document $d$ in the unit sphere embedding space, where $\mu \in \mathbb{R}^r$ is the mean vector and $\Sigma \in \mathbb{R}^{r \times r}$ is the covariance matrix. Let $\mathbf{d} \in \mathbb{R}^r$ be the embedding of document $d$, and let $\theta$ be the angle between $\mu$ and $\mathbf{d}$ such that $\cos(\theta) = \mu^\top \mathbf{d}$. Assume that both $\mu$ and $\mathbf{d}$ are unit vectors. Then, the following properties hold:
\begin{enumerate}
  \item The concentration inequality for the cosine similarity measure of $\mathbf{q}$ with $\mathbf{d}$ :
    \begin{small}
            \begin{equation}
    \label{query_doc}
    \mathbb{P}\left( \left| \mathbf{q}^\top \mathbf{d} - \cos(\theta) \right| \geq t \right) \leq 2 \exp\left( - \frac{t^2}{2 \mathbf{d}^\top \Sigma \mathbf{d} } \right).
    \end{equation}
    \end{small}
  \item The concentration inequality for the cosine similarity measure of $\mathbf{q}$ with  $\mu$ :
    \begin{equation}
    \label{query_mean}
    \mathbb{P}\left( \left| \mathbf{q}^\top \mu - 1 \right| \geq t \right) \leq 2 \exp\left( - \frac{t^2}{2 \mu^\top \Sigma \mu} \right).
    \end{equation}
  \item Non-Negativity of the Difference in Similarities:
  \begin{equation}
  \label{non-negative}
  \mathbf{q}\top \mu - \mathbf{q}\top \mathbf{d} = \mathbf{q}^\top \mu(1-\cos(\theta)) > 0.
    \end{equation}
\end{enumerate}
\end{theorem}
\begin{remark}
    The theorem provides a robust theoretical foundation for the QAEncoder's capabilities in computing the similarity between documents and queries.
These concentration inequalities in Equation \ref{query_doc} and Equation \ref{query_mean} show that cosine similarities between $\mathbf{q}$, $\mathbf{d}$, and $\mu$ are concentrated around their expected values.
Inequality \ref{non-negative} indicates that the similarity between the $\mathbf{q}$ and $\mu$ is always greater than between $\mathbf{q}$ and $\mathbf{d}$. This confirms that using the mean vector as a projection in QAEncoder better captures the semantic relationship between queries and documents. This theoretical result aligns with the experimental findings in Fig. \ref{fig:demo}, further validating QAEncoder's effectiveness.
\end{remark}

\subsubsection{Proof of Similarity Bounds}

\begin{proof}
Given the setup where $\mathbf{q} \sim \mathcal{N}(\mu, \Sigma)$ is an $r$-dimensional Gaussian random vector with mean $\mu$ and covariance $\Sigma$, and the angle between another unit vector $\mathbf{d}$ and $\mu$ is $\theta$.

The cosine of the angle between $\mathbf{q}$ and $\mathbf{d}$ is given by: $$ \cos(\phi_{qd}) = \mathbf{q}^\top \mathbf{d} .$$

Since $\mathbf{q}$ is Gaussian,  based on Lemma \ref{High-Dimensional_Gaussian_Distribution},the  inner product $\mathbf{q}^\top \mathbf{d}$ is a linear transformation of $\mathbf{q}$ and hence is a normal distribution with mean $\mu^\top \mathbf{d}$ and variance $\mathbf{d}^\top \Sigma \mathbf{d}$. Given that $\mu$ and $\mathbf{d}$ are unit vectors and the angle between $\mu$ and $\mathbf{d}$ is $\theta$, we have: $$ \mu^\top \mathbf{d} = \cos(\theta). $$

Thus, $\mathbf{q}^\top \mathbf{d}$ can be approximated as: $$ \mathbf{q}^\top \mathbf{d} \sim \mathcal{N}(\cos(\theta), \mathbf{d}^\top \Sigma \mathbf{d}). $$ 

The concentration inequality for the cosine value between $\mathbf{q}$ and $\mathbf{d}$ follows from Hoeffding's inequality for zero-mean sub-Gaussian random variables, which can be expressed as: $$ \mathbb{P}\left( \left| \mathbf{q}^\top \mathbf{d} - \cos(\theta) \right| \geq t \right) \leq 2 \exp\left( - \frac{t^2}{2 \mathbf{d}^\top \Sigma \mathbf{d} } \right) .$$

Similarly, the cosine of the angle between between $\mathbf{q}$ and $\mu$ can be expressed as: $$ \cos(\phi_{q\mu}) = \mathbf{q}^\top \mu .$$

Given that $\mathbf{q} \sim \mathcal{N}(\mu, \Sigma)$, based on Lemma \ref{High-Dimensional_Gaussian_Distribution},  the  inner product $\mathbf{q}^\top \mu$,
representing the cosine of the angle between $\mathbf{q}$ and $\mu$
is a linear transformation of a Gaussian random vector with mean: $$ \mathbb{E}[\mathbf{q}^\top \mu] = \mu^\top \mu = 1 ,$$ since $\mu$ is a unit vector, and variance: $$ \mathrm{Var}[\mathbf{q}^\top \mu] = \mu^\top \Sigma \mu .$$

Thus, $\mathbf{q}^\top \mu$ can be approximated as: $$ \mathbf{q}^\top \mu \sim \mathcal{N}(1, \mu^\top \Sigma \mu).$$ 

Applying the similar Hoeffding's inequality, the concentration inequality can be derived similarly: $$ \mathbb{P}\left( \left| \mathbf{q}^\top \mu - 1 \right| \geq t \right) \leq 2 \exp\left( - \frac{t^2}{2 \mu^\top \Sigma \mu} \right). $$

Notably, we observe that:$$\mathbf{q}^\top \mathbf{d} =(\mathbf{q}^\top \mu )(\mu^\top \mathbf{d})= (\mathbf{q}^\top \mu)\cos(\theta).$$
Therefore, comparing $\mathbf{q}^\top \mu$ and $\mathbf{q}^\top \mathbf{d}$, we find:

$$\mathbf{q}^\top \mu-\mathbf{q}^\top \mathbf{d}=\mathbf{q}^\top \mu(1-\cos(\theta)) .$$

Since $\mathbf{q} \neq \mathbf{d}$, it follows that $\cos(\theta) < 1$.
Moreover, $\mathbf{q}^\top \mu > 0 $ because $\mathbf{q}$ is a Gaussian random vector centered at $\mu$, which implies that $\mathbf{q}$ generally aligns positively with its mean $\mu$. Given that both $1 - \cos(\theta) > 0$ and $\mathbf{q}^\top \mu > 0$, we have:

$$
\mathbf{q}^\top \mu - \mathbf{q}^\top \mathbf{d} = \mathbf{q}^\top \mu (1 - \cos(\theta)) > 0 .
$$

\end{proof}

\subsubsection{Normality Test}
\label{Normality_Test}

To assess whether potential queries follow a Gaussian distribution in the embedding space,we employ two  statistical tests: the Chi-Squared Q-Q Plot and the Anderson-Darling Test. 

\paragraph{Chi-Squared Q-Q Plot Verification} We employ the Chi-Squared Q-Q Plot to assess whether the squared Mahalanobis distances~\ref{Squared_Mahalanobis_Distance} conform to the chi-squared distribution. By leveraging Lemma \ref{Chi-Squared Distribution}, We compare the observed $D^2$ values with the theoretical quantiles of the chi-squared distribution to assess the conformity of the data to a high-dimensional Gaussian model.
We conclude that close alignment of the sample points along the 45-degree reference line indicates support for the original hypothesis.

\begin{figure}
    \centering
    \includegraphics[width=\linewidth]{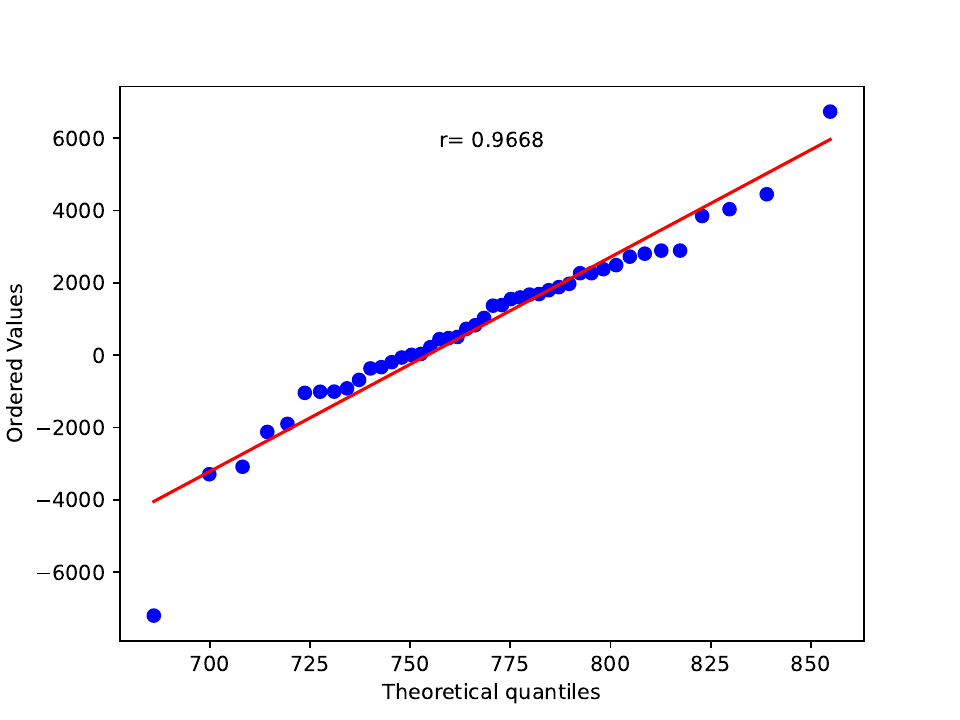}
    \caption{Q-Q plot against chi-squared distribution.}
    \label{Q_Q}
\end{figure}

Direct applications of high-dimensional normality tests, such as the Henze-Zirkler test, often lead to Type I errors in our case. E.g., testing on 768-dimensional normal samples revealed that Henze-Zirkler test demands high ratios of sample size to dimensionality, and it particularly susceptible to Type I errors when the sample size is not sufficiently large. Hence, we opted to perform uni-variate normality assessments on marginal distribution instead.
\paragraph{Anderson-Darling Test} To further evaluate the normality of marginal distributions, we conduct the Anderson-Darling test across all dimensions of the embeddings. The specific steps are as follows:
\begin{enumerate}
    \item Null Hypothesis:
    \begin{itemize}
        \item $H_0$: Each dimension's marginal distribution follows a Gaussian distribution.
        \item $H_1$: At least one dimension's marginal distribution does not follow a Gaussian distribution.
    \end{itemize}
    \item Testing Results:
    
Following the Anderson-Darling test, the results across all dimensions failed to reject $H_0$. This suggests that each dimension's marginal distribution can statistically be considered Gaussian, thereby supporting our hypothesis that potential queries conform to a high-dimensional Gaussian distribution.
\end{enumerate}

In summary, through the verification of squared Mahalanobis distances using the Chi-Squared Q-Q Plot and the evaluation of marginal distributions with the Anderson-Darling test, we validate the plausibility that potential queries conform to a high-dimensional Gaussian distribution within the embedding space.

\subsubsection{ Some properties of Gaussian distribution}
The statement and proof of our main results  contain some mathematical concepts. This section introduces these concepts, covering the fundamental  lemmas and definitions essential for understanding our analysis.

\begin{lemma} \citep{tong2012multivariate}
\label{High-Dimensional_Gaussian_Distribution}
Let $\mathbf{x}$ follow a multivariate Gaussian distribution:
$$
\mathbf{x} \sim \mathcal{N}(\mu, \Sigma),
$$
where $\mu$ is the mean vector and $\Sigma$ is the covariance matrix. For any linear transformation $A \mathbf{x} + \mathbf{b}$, the result is also multivariate normal:
$$
\mathbf{y} = A \mathbf{x} + \mathbf{b} \sim \mathcal{N}(A \mu + \mathbf{b}, A \Sigma A^\top).
$$
\end{lemma}
\remark{The lemma shows that multivariate Gaussians remain Gaussian under linear transformations.}

\begin{definition}(Squared Mahalanobis Distance \citep{mclachlan1999mahalanobis})
\label{Squared_Mahalanobis_Distance}
Assuming $\mathbf{x} \sim \mathcal{N}(\mu, \Sigma)$ in a high-dimensional Gaussian distribution, the squared Mahalanobis distance  $D^2$ is defined as:
$$D^2 = (\mathbf{x} - \mu)^T \Sigma^{-1} (\mathbf{x} - \mu).$$
\begin{remark}
This metric quantifies the distance between the observed value and the mean.
\end{remark}
\end{definition}

\begin{lemma}(Chi-Squared Distribution of Squared Mahalanobis Distance \citep{mardia2024multivariate}
)
\label{Chi-Squared Distribution}
    If  $\mathbf{x}$ is a $r$-dimensional Gaussian random variable, then the squared Mahalanobis distance $D^2$ follows a chi-squared distribution:
$$
D^2 \sim \chi^2_r,
$$
where $r$ denotes the dimensionality of the variable.
\end{lemma}

\begin{remark}
The property that the squared Mahalanobis distance follows a chi-squared distribution can be viewed as a form of dimensionality reduction.  By mapping a high-dimensional Gaussian variable to a scalar that encodes its deviation from the mean, adjusted for the covariance structure, this transformation reduces the complexity of the multivariate data while preserving key statistical properties in a single distance metric.
\end{remark}

\section{Query Generation Pipeline}
\label{app:pipeline}
\revise{
For general purposes, we mainly utilized GPT-4o-mini and the zero-shot prompt in Appendix \ref{prompts} for query generation on the latest datasets. For extra experiments on robustness, we also adopt Claude-3-5-Haiku-20241022, Qwen-Turbo-2024-11-01 and Deepseek-V3 as the query predictors. The temperature value is 0.95 and the frequency penalty is 0.1, making a balance between diversity and validity. That is, with a higher temperature value, the generated questions can become too divergent to be answered or supported by the documents.

This pipeline prompt is common and effective for automatic question generation \citep{zhang2025can, maity2024future}. We demonstrate the diversity of generated queries and the robustness of QAEncoder representations.

\subsection{The Diversity of Predicted Query}
\paragraph{Statistics Results}
\citet{zhang2025can} adopted almost the same prompt and generation process. It showed the diversity and quality of generated queries from six dimensions including \textit{question types, question length, context coverage, answerability, uncommonness, and answer length}. We excerpt the table of question types, showing the diversity of generated queries.

\begin{table}[H]
    \centering
    \resizebox{\columnwidth}{!}{
    \begin{tabular}{|l|l|l|l|l|}
    \hline
        ~ & TriviaQA & HotpotQA & Llama & GPT \\ \hline
        T1 Identity/Attribution & 34.2 & 39.7 & 7.5 & 15.7 \\ \hline
        T2 General Knowledge & 34.5 & 15.0 & 7.3 & 12.3 \\ \hline
        T3 Location & 12.2 & 14.3 & 3.1 & 4.3 \\ \hline
        T4 Classification/Categorization & 4.3 & 2.7 & 2.1 & 1.5 \\ \hline
        T5 Specific Fact/Figure & 10.5 & 9.5 & 18.7 & 24.1 \\ \hline
        T6 Comparison/Selection & 0.1 & 6.7 & 1.0 & 0.6 \\ \hline
        T7 Verification/Affirmation & 0.1 & 6.5 & 0.1 & 0.2 \\ \hline
        T8 Descriptive/Characterization & 3.0 & 1.5 & 43.8 & 28.7 \\ \hline
        T9 Event/Outcome & 0.2 & 0.8 & 14.7 & 10.1 \\ \hline
        T10 Sequential/Ordering/Causation & 0.9 & 3.2 & 1.7 & 2.4 \\ \hline
        Others & 0.0 & 0.0 & 0.0 & 0.1 \\ \hline
    \end{tabular}
    }
    \caption{Percentage of different question types across different datasets}
\end{table}

\paragraph{Case Study}
To provide a general feeling of diversity, we show the predicted queries by GPT and Claude. For two constructed news benchmarks, the test query is sampled without replacement from GPT's results; for BEIR, the test queries are from original benchmarks.
}
\begin{tcolorbox}[colback=gray!10, colframe=black!75, title=Case Study of Diversity]
\begin{lstlisting}[basicstyle=\footnotesize\ttfamily,breaklines=true,aboveskip=\medskipamount,belowskip=\medskipamount,showstringspaces=false,xleftmargin=0pt]
- Document: On July 20, 1969, Neil Armstrong landed on the moon to accomplish the Apollo 11 mission, achieved by NASA's planning and the Saturn V rocket.
- GPT: ['1. What significant event took place on July 20, 1969?', '2. Who was the first person to set foot on the moon?', '3. Which organization was responsible for planning the Apollo 11 mission?', '4. What was the name of the rocket used to accomplish the Apollo 11 mission?', '5. What achievement was accomplished by Neil Armstrong during the Apollo 11 mission?']
- Claude: ['1. On what date did Neil Armstrong land on the moon?', '2. What was the name of the mission during which Neil Armstrong landed on the moon?', '3. Which organization planned the moon landing?', '4. What type of rocket was used to achieve the moon landing?', '5. Who was the astronaut who landed on the moon during the Apollo 11 mission?']
\end{lstlisting}
\end{tcolorbox}

\section{Cost-Effectiveness}
\label{append_cost}
Briefly, QAEncoder is cost-effective and efficient.

\begin{figure*}
    \centering
    \includegraphics[width=0.45\linewidth]{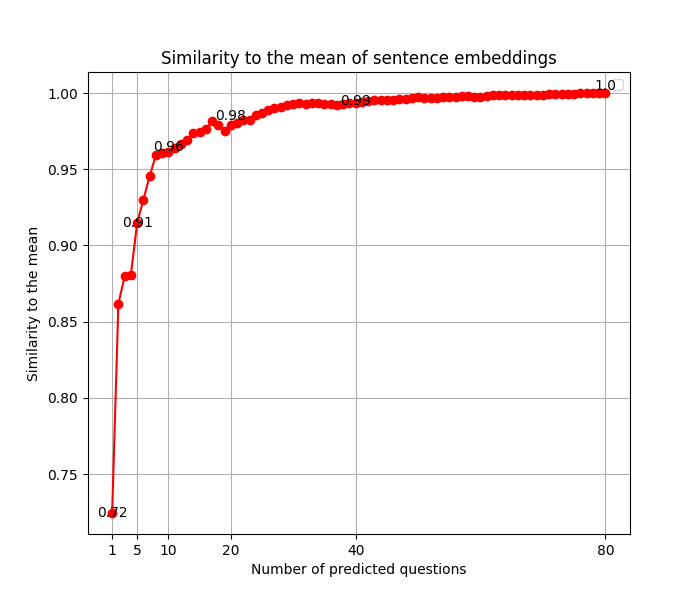}
    \caption{
The convergence speed for Monte Carlo estimation of QAE\textsubscript{base}, $n$ denotes the number of prediction queries. For documents with a length greater than 150 words from MSMARCO datasets, generating 10 queries exhibits a similarity score of 0.96 compared to generating 80 queries. Besides, the document fingerprint strategies introduces the hyperparameter $\alpha$ to QAE\textsubscript{base}, further reducing the variance.
    }
    \label{fig:monte_carol}
\end{figure*}

\begin{itemize}
    \item Firstly, the cost mainly comes from the query generation process during indexing phase. As shown in Fig. \ref{fig:monte_carol}, 10 queries with an average length of 6-15 tokens are typically sufficient for ordinary documents.
    \item Secondly, GPT-4o-mini is currently the cheapest generative model by OpenAI, priced at \$0.075 per million input tokens and \$0.300 per million output tokens. The most recent Qwen2.5-Turbo by Alibaba is even 3.6 times cheaper than GPT-4o-mini, which outperforms GPT-4 on long-context and mirrors GPT-4o-mini on short-sequence \citep{Qwen25Turbo}. Hence, about 0.1 million documents can be processed within 1 dollar via API call ($\frac{1\mathrm{M}*3.6}{0.3*(10*10)} \approx 0.1\mathrm{M}$). The prices will continually decrease as AI develops.
    \item Thirdly, the initial query-centric work for sparse retrievers, DocT5Query, demonstrates T5-base with 0.2B parameters is sufficient for query generation, while no improvement are gained with larger models \citep{docT5query}. We also confirmed Qwen2.5-0.5B-Instruct as a good query generator in our business implementation. We choose GPT-4o-mini just given its out-of-the-box and comprehensively multilingual support for academic research.
    \item Finally, in our data flow settings (batch size=1), 2-3 documents can be processed per second with Qwen2.5-0.5B-Instruct + vLLM + BF16 on a single NVIDIA A100 80GB \cite{Qwen25SpeedBenchmark}. For batch processing with higher GPU utility, DocT5Query reports sampling 5 queries per document for 8.8M MSMARCO documents requires approximately 40 hours on a single Google TPU v3, costing only \$96 USD (40 hours × \$2.40 USD/hour) in 2019 \citep{docT5query}.
\end{itemize}

\section{Dataset Details}
\label{dataset}
\begin{itemize}
    \item The BEIR benchmark is a meticulously curated collection of 19 datasets, designed to comprehensively evaluate the generalization capabilities of information retrieval (IR) models across a wide range of heterogeneous tasks.
    Among the 19 datasets, 15 are publicly available and are used for evaluation in our experiments. The 15 datasets selected from BEIR encompass diverse domains: MSMARCO \citep{MSMARCO}, TREC-COVID(TRECC.) \citep{voorhees2021trec}, NFCorpus \citep{boteva2016full}, Natural Questions (NQ) \citep{Naturalquestions}, HotpotQA \citep{yang2018hotpotqa}, FiQA18 \citep{maia201818}, ArguAna \citep{wachsmuth2018retrieval}, Touche20 \citep{bondarenko2022overview}, CQADupStack(CQADup.) \citep{hoogeveen2015cqadupstack}, Quora, DBPedia \citep{hasibi2017dbpedia}, SciDocs \citep{cohan2020specter}, Fever \citep{thorne2018fever}, Climate-Fever(ClimateFe.) \citep{diggelmann2020climate}, and SciFact \citep{wadden2020fact}.

    \item %
    CRUD-RAG is a benchmark specifically designed for evaluating RAG systems, incorporating the latest high-quality news data that were not included in the training phase of the language models. It comprises more than 80K news articles sourced from prominent Chinese news websites, all published after July 2023.
    From the set of queries generated by GPT-4o-mini for each document, we randomly sample one to serve as the test query. The original news is designated as the evidence documents for recall evaluation, ensuring queries are associated with exactly relevant documents.\footnote{LlamaIndex adopts the same common practice and provides templated workflow, i.e. generating queries for recall and rerank evaluation. See \href{https://www.llamaindex.ai/blog/boosting-rag-picking-the-best-embedding-reranker-models-42d079022e83}{the website}.}

    \item FIGNEWS is a multilingual news post dataset designed to examine bias and propaganda within news articles across different languages. It consists of 15,000 publicly available news posts collected from verified blue-check accounts between October 7, 2023, and January 31, 2024. The dataset includes posts in five languages—English, Arabic, Hebrew, French, and Hindi—distributed evenly across 15 batches, each containing 1,000 posts. Each batch consists of 200 posts for each language. Similar to CRUD-RAG, we randomly sample one predicted query generated by GPT-4o-mini for each document and use the original news as the evidence documents for recall evaluation.
    
\end{itemize}

\section{Metric Details}
\label{metric}
\begin{itemize}
    \item Mean Reciprocal Rank (MRR):
    Mean Reciprocal Rank: is a statistic measure used to evaluate the effectiveness of a retrieval system by calculating the reciprocal of the rank at which the first relevant result appears. The mathematical formulation is:
    
    $$ \text{MRR} = \frac{1}{|Q|} \sum_{i=1}^{|Q|} \frac{1}{\text{rank}_i} $$
    
    Where: - $ |Q| $ is the number of queries. - $ \text{rank}_i $ is the rank position of the first relevant document for the $ i $-th query.

    \item Normalized Discounted Cumulative Gain (NDCG):
    Normalized Discounted Cumulative Gain is a measure of ranking quality that takes into account the positions of the relevant documents. It is based on the concept of discounting the relevance of documents based on their position in the result list. The mathematical formulation is:
    
    $$ \text{NDCG} = \frac{DCG_p}{IDCG_p}, $$
    
    Where: - $ DCG_p $ is the Discounted Cumulative Gain at position $ p $. - $ IDCG_p $ is the Ideal Discounted Cumulative Gain at position $ p $, which is the DCG score of the perfect ranking.
    
    The Discounted Cumulative Gain at position $ p $ is given by:
    
    $$ DCG_p = \sum_{i=1}^{p} \frac{2^{\text{rel}_i} - 1}{\log_2(i + 1)}, $$
    
    Where: - $ \text{rel}_i $ is the relevance score of the document at rank $ i $.
    
    The Ideal Discounted Cumulative Gain $ IDCG_p $ is computed in the same way as $ DCG_p $, except that the documents are ideally sorted by relevance.
\end{itemize}

\section{Baseline Details}
\label{baseline}
\begin{itemize}
\item BM25 \citep{BM25} is the traditional lexical retriever based on term relevance and frequency, regarded as the most popular variation of TF-IDF.
\item DocT5Query \citep{docT5query} appends generated queries to the document before building the inverted index of BM25.
\item BGE models \citep{BGE}by BAAI, Jina models \citep{gunther2023jina} by Jina AI, E5 models \citep{wang2023improving}  by Microsoft, GTE models \citep{zhang2024mgte}  by Alibaba-NLP are the most advanced embedding models, featuring multilingual understanding and task-specific instruction tuning capabilities. We choose both vanilla encoders and QA-specific instruction-tuned encoders for test.
\item Contriever models \citep{Contriever} are developed by Facebook Research, including contriever, mcontriever, contriever-msmacro and mcontriever-msmacro. mcontriever serves as the multilingual version of contriever. contriever-msmacro and mcontriever-msmacro are further fine-tuned on the MS-MACRO dataset for bridging the document-query gap.
\item BCEmbedding models \citep{youdao_bcembedding_2023}, developed by NetEase Youdao, are bilingual and crosslingual embedding models in English and Chinese. BCEmbedding serves as the cornerstone of Youdao's RAG-based QA system, QAnything, an open-source project widely integrated in commercial products like Youdao Speed Reading and Youdao Translation. We choose bce-embedding-base for test.
\item Text2Vec models \citep{Text2vec} is a popular open-source project that implements Word2Vec \citep{Word2Vec}, RankBM25, BERT \citep{Bert}, Sentence-BERT \citep{Sentence-BERT}, CoSENT and other text representation models. We test its most prominent model, text2vec-base-multilingual, which supports multiple languages, including German, English, Spanish, French, Italian, Dutch, Polish, Portuguese, Russian, and Chinese.
\item 
DPR models \citep{karpukhin2020dense} by Facebook adopt a bi-encoder architecture. DPR models fine-tuned BERT on pairs of questions and passages without additional pretraining, achieving superior performance compared to traditional methods like BM25.

\item  
\textcolor{black}{
RePAQ models \citep{lewis2021paq} are dense retrievers trained on PAQ \citep{lewis2021paq}, 
a large-scale synthetic corpus of question–answer pairs generated from web texts. We evaluate retriever-multi-base-256, a multilingual variant designed for efficient passage retrieval across multiple languages.}
\item 
\textcolor{black}{Quora DistilBERT models \citep{Sentence-BERT}, released by the Sentence-Transformers project, are sentence embedding models based on DistilBERT and fine-tuned on the Quora Duplicate Questions dataset \citep{thakur2021beir} for semantic similarity and duplicate question detection. We evaluate both quora-distilbert-base and quora-distilbert-multilingual, the latter being a multilingual variant trained on parallel corpora covering over 50 languages.}

\item \textcolor{black}{Training-based approaches mainly include two types: fine-tuning on QA datasets (domain adaptation) and fine-tuning on multi-task instruction datasets.}
For fine-tuning on QA datasets, we choose mcontriever-msmarco for test, an enhanced variant of the mcontriever model that has been fine-tuned on the MSMARCO.
The second category involves fine-tuning models on multi-task instruction datasets, where distinct prompt prefixes are appended to the input text, enabling the model to effectively differentiate between various tasks. In this category, we test the multilingual-e5-large-instruct \citep{e52024multilingual} developed by Microsoft, which leverages synthetic instruction data \citep{wang2023improving} for fine-tuning.
\textcolor{black}{To unveil the catastrophic forgetting issue of training-based methods, we incorporate GPL \cite{wang2021gpl}, which predicted queries, mines hard negative samples, and distills the re-ranker for unsupervised domain adaptation. The method utilize pseudo-queries generated from the target domain as supervision for contrastive learning.
Following the original settings, we train the GPL models for 140,000 steps with a batch size of 32. }

\item Document-centric methods instruct LLMs to generate a pseudo-document for each query. The pseudo-document aims to capture relevant information but does not correspond to a real document and may contain inaccuracies and hallucinations. Subsequently,the pseudo-document is encoded, and its embedding is utilized to retrieve similar real documents based on vector similarity.
\textcolor{black}{We choose HyDE \citep{HyDE}, Query2Doc \citep{Query2doc}, and QA-RAG \citep{QA-RAG} for test. 
Hyde generates multiple pseudo-documents and fuses their embeddings by mean pooling for retrieval.
Similarly, Query2Doc concatenates the query with the generated pseudo-document and performs dense retrieval using the embedding of the combined text.
QA-RAG \citep{QA-RAG} enhances retrieval through a two-way mechanism that utilizes both user query and pseudo-documents for respective retrieval.}

\end{itemize}

\section{Instruction Templates}
\label{prompts}
We employ specialized prompts to instruct GPT-4o-mini as the question and pseudo-document generator respectively. Only the English prompts are presented due to LaTeX compilation issues with non-English languages. For BEIR benchmarks, declarative words can serve as valid user queries. We adopt the few-shot in-context learning, which enables the model to generate declarative queries that not only resemble natural inputs but also serve a role analogous to document fingerprints.

\begin{tcolorbox}[colback=gray!10, colframe=black!75, title=Question Generator Prompt]
\begin{lstlisting}[basicstyle=\footnotesize\ttfamily,breaklines=true,aboveskip=\medskipamount,belowskip=\medskipamount,showstringspaces=false,xleftmargin=0pt]
Context information is below. 
---------------------
[Document]
---------------------
Given the context information and not prior knowledge, generate only questions based on the below query.
You are a Teacher/Professor. Your task is to setup [Number of Questions] questions for an upcoming quiz/examination. The questions should be diverse in nature across the document. Restrict the questions to the information provided, and avoid ambiguous references. 

Output Format:
```json
[
    "1. question",
    "2. question",
    ...
]
```
\end{lstlisting}
\end{tcolorbox}

\begin{tcolorbox}[colback=gray!10, colframe=black!75, title=Pseudo-Document Generator Prompt]
\begin{lstlisting}[basicstyle=\footnotesize\ttfamily,breaklines=true,aboveskip=\medskipamount,belowskip=\medskipamount,showstringspaces=false,xleftmargin=0pt]
Please write a passage to answer the question.
Question: [Question]
Output Format:
```json
{
    "passage": ""
}
```
\end{lstlisting}
\end{tcolorbox}

\section{Hyperparameter Selection}
\label{hyperparameter_search}
As shown in the main body, $\text{QAE}_{\text{emb}}$ maintains competitive performance with single hyperparameter. Hence, $\text{QAE}_{\text{emb}}$ is recommended for accelerating HP search. 

\begin{itemize}
    \item Firstly, we believe that the optimal hyperparameters are primarily influenced by the inherent characteristics of the embedding model, i.e. the geometric property of embedding space. Therefore, a one-turn search should be sufficient for a given embedding model. That's why we optimize hyperparameters simultaneously across multiple datasets.
    \item Secondly, the one-turn search can also be accelerated under our framework. Indeed, as Fig. \ref{fig:Ablation_alpha_naive} shows, the performance of $\text{QAE}_{\text{emb}}$ empirically follows a consistent trend across various models and datasets: it initially rises and then falls as $\alpha$ increases, peaking between 0.3 and 0.6. This unimodal phenomenon enables ternary search  with logarithmic trails rather than brute-force search.
    \item Finally, the property of datasets also slightly influences the optimal hyperparameters. Specifically, the optimal $\alpha$ for classical datasets is marginally lower than that for latest datasets (refer to Tables \ref{full_classic}
 and \ref{full_latest}  for details). Therefore, selecting the optimal $\alpha$ based on classical datasets represents a cautious and robust strategy, ensuring consistent improvement across both classical and latest datasets.
\end{itemize}

\section{Challenges of Existing Methods}
\subsection{Challenges of QAE\textsubscript{naive}}
\label{naive_challenges}
\begin{itemize}
    \item[\textbf{C1.}] \textbf{Expanded Index Size.}
    Storing all QA pairs significantly increases the index size, leading to a substantial expansion in storage requirements, especially problematic for large-scale corpora.
    \item[\textbf{C2.}] \textbf{Prolonged Retrieval Times.}
    The index expansion also results in extended retrieval times. For dense retrievers, the expanded index size can result in linearly increased search time in both exhaustive and non-exhaustive search \citep{douze2024faiss} and hurt the recall performance in non-exhaustive case \citep{zhao2023towards}.
    \item[\textbf{C3.}] \textbf{Limited Query Handling.}
    Although storing QA pairs individually can address predicted queries, this approach lacks robustness when confronted with the wide-ranging and diverse nature of potential queries \citep{alting2020evaluating}.
\end{itemize}
\subsection{Challenges of Document-centric Methods}
\label{document_challenges}
Document-centric methods such as HyDE and Query2Doc suffer from not only computation overhead at inference time but also hallucination of pseudo-document generation. Specifically, for user queries, these methods generate pseudo-documents as retrieval queries. However, pseudo-documents suffer from hallucination, especially with rapidly updated knowledge. 
A concrete example hallucination can be:
\begin{tcolorbox}[colback=gray!10, colframe=black!75, title=Pseudo-Document Generator Prompt]
\begin{lstlisting}[basicstyle=\footnotesize\ttfamily,breaklines=true,aboveskip=\medskipamount,belowskip=\medskipamount,showstringspaces=false,xleftmargin=0pt]
- User Query: Who won the 2024 Abel Prize?
- Pseudo Document (Hallucinatory Retrieval Query): The 2024 Abel Prize was awarded to Abel for proving that it is impossible to solve the general equation of the fifth degree using radicals.
- Retrieved Document: Abel's most famous single result is the first complete proof demonstrating the impossibility of solving the general quintic equation in radicals. This question was one of the outstanding open problems of his day, and had been unresolved for over 250 years.[2] He was also an innovator in the field of elliptic functions and the discoverer of Abelian functions.
- LLM Answer: Abel
- Groundtruth: Michel Talagrand
\end{lstlisting}
\end{tcolorbox}

\section{More Figures and Tables}
\label{full_tables}

\begin{figure*}
    \centering
    \includegraphics[width=0.45\linewidth]{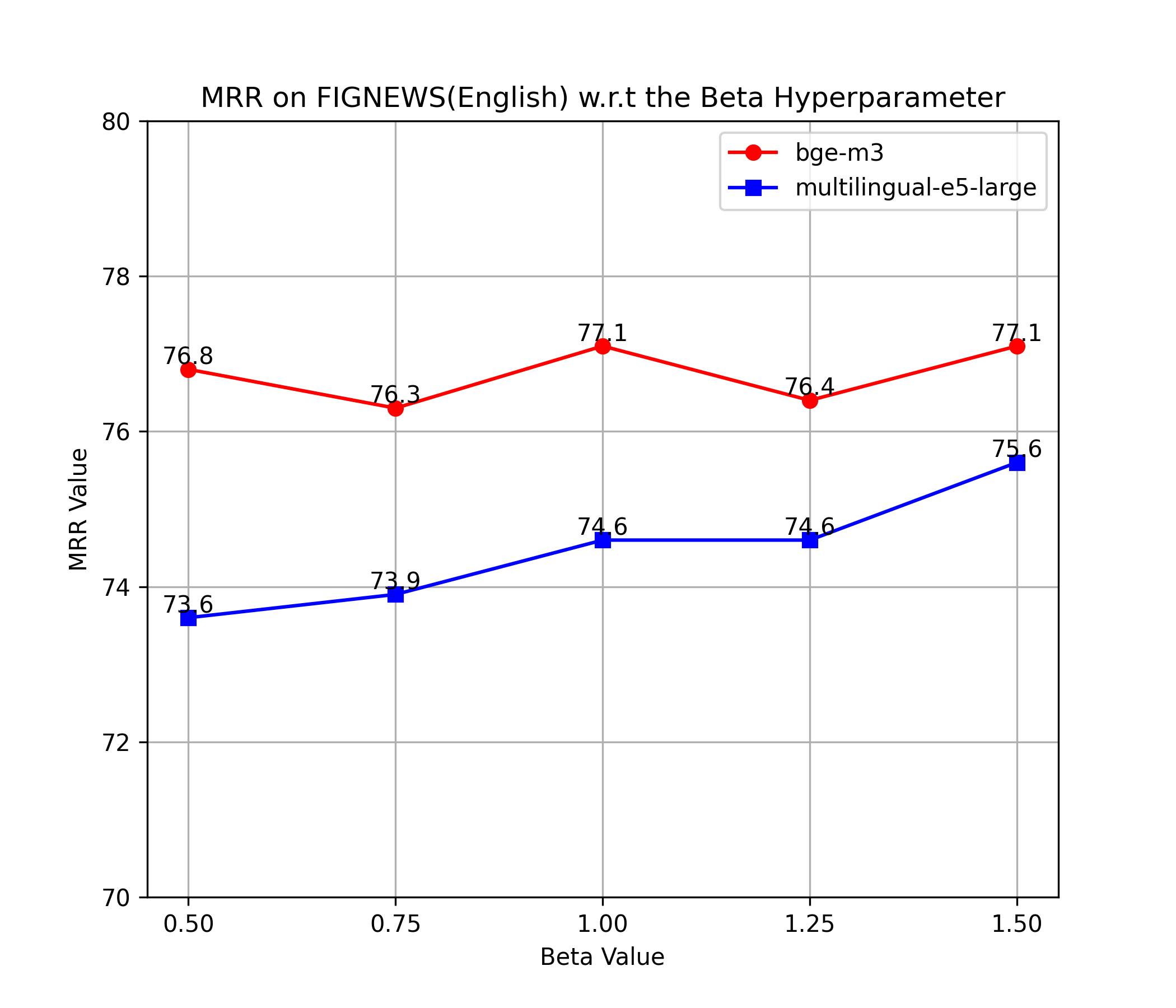}
    \caption{Ablation on $\beta$ hyperparameter for QAE\textsubscript{txt} on FIGNEWS(English) dataset.}
    \label{fig:ablation_txt}
\end{figure*}

\begin{figure*}
    \centering
    \includegraphics[width=\linewidth]{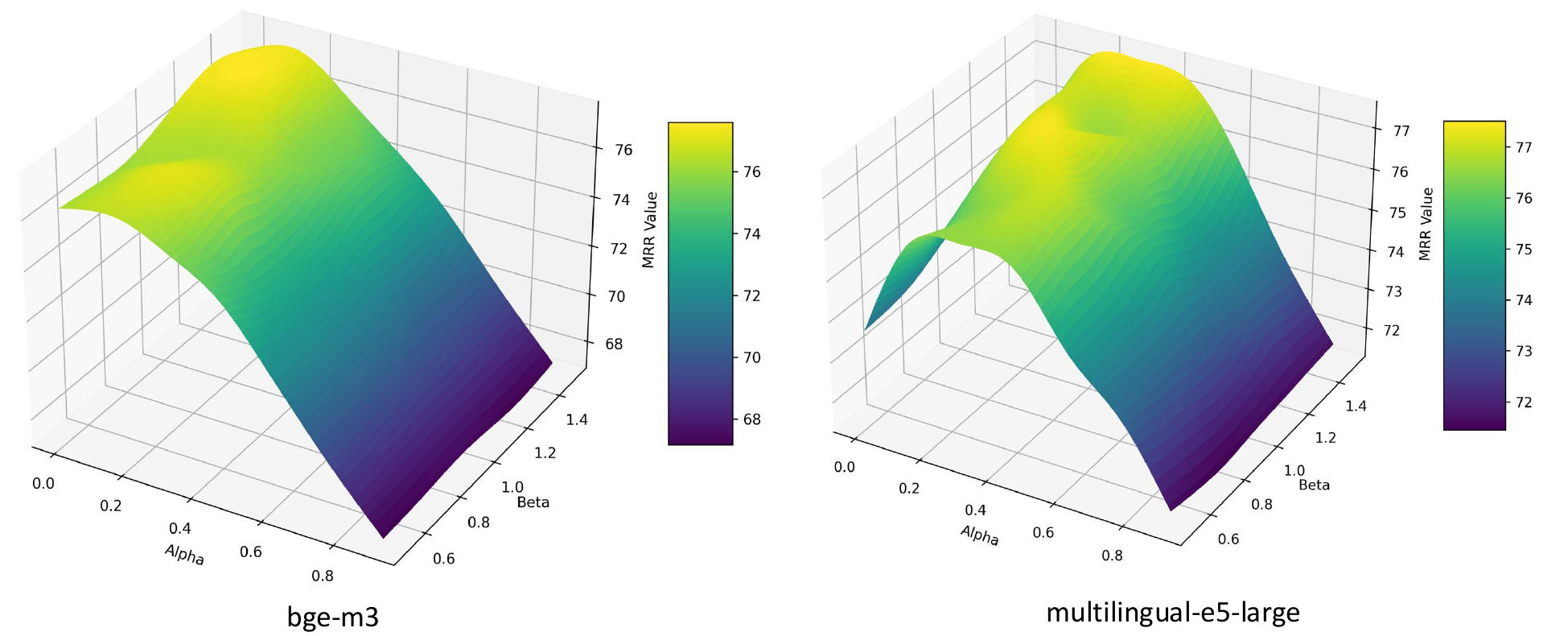}
    \caption{Ablation on $\alpha$ and $\beta$ hyperparameters for QAE\textsubscript{hyb} on FIGNEWS(English) dataset.}
    \label{fig:ablation_hyb}
\end{figure*}

\begin{table*}[htbp]

  \centering
  \vspace{10pt}
  \resizebox{\textwidth}{!}{
  \begin{tabular}{ccccccccccccccccc}
  \midrule
    Model & Method & ArguAna & ClimateF. & CQADups. & DBPedia & FEVER & FiQA18 & HotpotQA & MSMARCO & NFCorpus & NQ    & Quora & SciDocs & SciFact & Touche20 & TRECC. \\
    \midrule
    \multicolumn{17}{c}{\textbf{Sparse}} \\
    \midrule
    BM25  & -     & 31.5  & \textbf{21.3} & 29.9  & 31.3  & \textbf{75.3} & 23.6  & \textbf{60.3} & 22.8  & \textbf{32.5} & 32.9  & 78.9  & 15.8  & 66.5  & \textbf{36.7} & 65.6 \\
    DocT5Query & -     & \textbf{34.9} & 20.1  & \textbf{32.5} & \textbf{33.1} & 71.4  & \textbf{29.1} & 58.0  & \textbf{33.8} & \textbf{32.8} & \textbf{39.9} & \textbf{80.2} & \textbf{16.2} & \textbf{67.5} & 34.7  & \textbf{71.3} \\
    \midrule
    \multicolumn{17}{c}{\textbf{Dense}} \\
    \midrule
    \multirow{2}[1]{*}{dpr} & -     & 17.5  & 14.8  & 15.3  & 26.3  & 56.2  & 11.2  & 39.1  & 17.7  & 18.9  & 47.4  & 24.8  & 7.7   & 31.8  & 13.1  & 33.2 \\
          & QAE\textsubscript{emb}, $\alpha=0.45$ & \textbf{29.6} & \textbf{27.6} & \textbf{22.3} & \textbf{32.4} & \textbf{70.9} & \textbf{20.3} & \textbf{43.4} & \textbf{25.4} & \textbf{23.9} & \textbf{52.7} & \textbf{28.9} & \textbf{13.5} & \textbf{40.4} & \textbf{22.9} & \textbf{48.0} \\
          \midrule
    \multirow{2}[0]{*}{contriever} & -     & 37.9  & 15.5  & 28.4  & 29.2  & 68.2  & 24.5  & 48.1  & 20.6  & 31.7  & 25.4  & 83.5  & 14.9  & 64.9  & 19.3  & 27.4 \\
          & QAE\textsubscript{emb}, $\alpha=0.45$ & \textbf{47.1} & \textbf{20.2} & \textbf{33.5} & \textbf{33.7} & \textbf{73.1} & \textbf{30.4} & \textbf{50.3} & \textbf{26.1} & \textbf{37.9} & \textbf{28.1} & \textbf{86.2} & \textbf{17.3} & \textbf{70.2} & \textbf{25.2} & \textbf{45.2} \\
          \midrule
    \multirow{2}[0]{*}{contriever-msmarco} & -     & 44.6  & 23.7  & 34.5  & 41.3  & 75.8  & 32.9  & 63.8  & \textbf{40.7} & 32.8  & 49.8  & 86.5  & 16.5  & 67.7  & 20.4  & 59.6 \\
          & QAE\textsubscript{hyb}, $\alpha=0.3,\;\beta=0.75$ & \textbf{53.9} & \textbf{25.5} & \textbf{38.2} & \textbf{44.9} & \textbf{82.0} & \textbf{37.9} & \textbf{65.3} & 39.9  & \textbf{37.6} & \textbf{52.7} & \textbf{89.9} & \textbf{18.7} & \textbf{73.6} & \textbf{26.3} & \textbf{70.6} \\
          \midrule
    \multirow{2}[0]{*}{bge-large-en-v1.5} & -     & 63.5  & 36.6  & 42.2  & 44.1  & 87.2  & 45.0  & \textbf{74.1} & \textbf{42.5} & 38.1  & 55.0  & 89.1  & 22.6  & 74.6  & 24.8  & 74.8 \\
          & QAE\textsubscript{hyb}, $\alpha=0.15,\;\beta=0.5$ & \textbf{68.8} & \textbf{38.2} & \textbf{45.6} & \textbf{44.8} & \textbf{91.5} & \textbf{48.2} & 73.9  & 41.2  & \textbf{41.7} & \textbf{56.4} & \textbf{90.1} & \textbf{25.6} & \textbf{78.9} & \textbf{28.1} & \textbf{78.2} \\
          \midrule
    \multirow{2}[0]{*}{multilingual-e5-large} & -     & 54.4  & 25.7  & 39.7  & 41.3  & \textbf{82.8} & 43.8  & \textbf{71.2} & \textbf{43.7} & 34.0  & 64.1  & \textbf{88.2} & 17.5  & 70.4  & 23.1  & 71.2 \\
          & QAE\textsubscript{hyb}, $\alpha=0.15,\;\beta=1.5$ & \textbf{61.1} & \textbf{28.4} & \textbf{44.3} & \textbf{42.9} & 82.1  & \textbf{45.1} & 69.9  & 43.0  & \textbf{35.5} & \textbf{65.3} & 88.0  & \textbf{20.8} & \textbf{73.9} & \textbf{26.3} & \textbf{75.1} \\
          \midrule
    \multirow{2}[0]{*}{e5-large-v2} & -     & 46.4  & 22.2  & 37.9  & 44.0  & 82.8  & 41.1  & 73.1  & \textbf{43.5} & 37.1  & 63.4  & 86.8  & 20.5  & 72.2  & 20.7  & 66.5 \\
          & QAE\textsubscript{hyb}, $\alpha=0.3,\;\beta=1.0$ & \textbf{55.1} & \textbf{25.3} & \textbf{41.2} & \textbf{45.5} & \textbf{86.5} & \textbf{43.4} & \textbf{73.9} & 42.8  & \textbf{39.8} & \textbf{64.8} & \textbf{89.5} & \textbf{23.3} & \textbf{75.3} & \textbf{23.8} & \textbf{74.2} \\
          \midrule
    \multirow{2}[0]{*}{gte-base-en-v1.5} & -     & 63.5  & 40.4  & 39.5  & \textbf{39.9} & \textbf{94.8} & 48.7  & \textbf{67.8} & \textbf{42.6} & 35.9  & \textbf{53.0} & \textbf{88.4} & 21.9  & 76.8  & 25.2  & 73.1 \\
          & QAE\textsubscript{hyb}, $\alpha=0.3,\;\beta=0.5$ & \textbf{68.2} & \textbf{43.2} & \textbf{43.7} & 39.6  & 94.2  & \textbf{51.6} & 66.7  & 41.9  & \textbf{38.7} & 52.3  & 88.2  & \textbf{24.8} & \textbf{80.3} & \textbf{29.3} & \textbf{77.5} \\
          \midrule
    \multirow{2}[0]{*}{jina-embeddings-v2-small-en} & -     & 46.7  & 24.0  & 38.0  & 32.7  & 68.0  & 33.4  & \textbf{56.5} & 37.3  & 30.4  & \textbf{51.6} & 87.2  & 18.6  & 63.9  & 23.5  & 65.2 \\
          & QAE\textsubscript{hyb}, $\alpha=0.15,\;\beta=0.5$ & \textbf{55.3} & \textbf{27.8} & \textbf{42.3} & \textbf{35.6} & \textbf{74.3} & \textbf{36.5} & 55.7  & \textbf{39.5} & \textbf{32.7} & 51.1  & \textbf{89.9} & \textbf{21.2} & \textbf{67.2} & \textbf{27.2} & \textbf{73.6} \\
          \midrule
    \end{tabular}%

    }

  \caption{\textcolor{black}{Complete retrieval performance across fifteen classical  datasets in BEIR (NDCG@10).  Higher is better, with the best one is bolded. Hyperparameters including QAEncoder variants and weight terms $\alpha,\;\beta$ are optimized simultaneously. `-' denotes default or null values.}}
  \label{full_classic}
\end{table*}

\begin{table*}[htbp]
  \centering
  \vspace{10pt}
  \resizebox{\textwidth}{!}{

\begin{tabular}{cccccccccccccc}
\toprule
\multirow{2}[4]{*}{Model} & \multirow{2}[4]{*}{Method} & \multicolumn{2}{c}{FIGNEWS(English)} & \multicolumn{2}{c}{FIGNEWS(Arabic)} & \multicolumn{2}{c}{CRUD-RAG(Chinese)} & \multicolumn{2}{c}{FIGNEWS(French)} & \multicolumn{2}{c}{FIGNEWS(Hindi)} & \multicolumn{2}{c}{FIGNEWS(Hebrew)} \\
\cmidrule{3-14}      &       & \multicolumn{1}{l}{MRR@10} & \multicolumn{1}{l}{NDCG@10} & \multicolumn{1}{l}{MRR@10} & \multicolumn{1}{l}{NDCG@10} & \multicolumn{1}{l}{MRR@10} & \multicolumn{1}{l}{NDCG@10} & \multicolumn{1}{l}{MRR@10} & \multicolumn{1}{l}{NDCG@10} & \multicolumn{1}{l}{MRR@10} & \multicolumn{1}{l}{NDCG@10} & \multicolumn{1}{l}{MRR@10} & \multicolumn{1}{l}{NDCG@10} \\
\midrule
\multirow{2}[2]{*}{bge-m3} & -     & 74.4  & 78.7  & 77.8  & 80.9  & 47.5  & 48.6  & 73.5  & 77.4  & 58.6  & 64.4  & 78.7  & 81.4 \\
      & QAE\textsubscript{txt}, $\beta=1.5$ & \textbf{77.2} & \textbf{81} & \textbf{80.2} & \textbf{83.1} & \textbf{51.4} & \textbf{52.5} & \textbf{76.9} & \textbf{80.3} & \textbf{62.7} & \textbf{67.8} & \textbf{80} & \textbf{82.8} \\
\midrule
\multirow{2}[2]{*}{multilingual-e5-small} & -     & 71    & 75.1  & 74.1  & 77.4  & 44.6  & 46.0  & 66.8  & 70.4  & 52.5  & 57.8  & 72.9  & 76.5 \\
      & QAE\textsubscript{hyb}, $\alpha=0.3,\;\beta=0.5$ & \textbf{74.6} & \textbf{78.5} & \textbf{78.9} & \textbf{81.6} & \textbf{50.6} & \textbf{51.6} & \textbf{74.2} & \textbf{77.9} & \textbf{59.6} & \textbf{64.6} & \textbf{77.5} & \textbf{80.4} \\
\midrule
\multirow{2}[2]{*}{multilingual-e5-base} & -     & 74.8  & 78.1  & 72.3  & 76    & 47.0  & 48.2  & 71.2  & 75.1  & 57.8  & 62.9  & 72.6  & 75.8 \\
      & QAE\textsubscript{emb}, $\alpha=0.3$ & \textbf{77.6} & \textbf{81.3} & \textbf{77.2} & \textbf{80.3} & \textbf{51.2} & \textbf{52.3} & \textbf{76.7} & \textbf{80} & \textbf{61.5} & \textbf{66.5} & \textbf{77.7} & \textbf{80.5} \\
\midrule
\multirow{2}[2]{*}{multilingual-e5-large} & -     & 73.9  & 77.8  & 76.7  & 80.2  & 46.9  & 48.3  & 70.6  & 74.5  & 53    & 59.2  & 73.9  & 77.4 \\
      & QAE\textsubscript{hyb}, $\alpha=0.15,\;\beta=1.25$ & \textbf{77.1} & \textbf{80.6} & \textbf{82.2} & \textbf{85.1} & \textbf{51.5} & \textbf{52.7} & \textbf{77.4} & \textbf{80.9} & \textbf{60.6} & \textbf{65.9} & \textbf{77.7} & \textbf{81} \\
\midrule
\multirow{2}[2]{*}{gte-multilingual-base} & -     & 65.5  & 70.4  & 73.4  & 76.8  & 45.3  & 46.8  & 63    & 67.4  & 52.4  & 58.6  & 66    & 69.6 \\
      & QAE\textsubscript{hyb}, $\alpha=0.15,\;\beta=1.5$ & \textbf{75.5} & \textbf{79.5} & \textbf{76.2} & \textbf{79.1} & \textbf{49.4} & \textbf{51.0} & \textbf{66.9} & \textbf{71.6} & \textbf{56} & \textbf{61.9} & \textbf{74} & \textbf{77.6} \\
\midrule
\multirow{2}[2]{*}{mcontriever} & -     & 32.9  & 36.7  & 40.3  & 44.7  & 39.2  & 41.6  & 35    & 39.3  & 27    & 31.8  & 49.5  & 54.4 \\
      & QAE\textsubscript{hyb}, $\alpha=0.45,\;\beta=1.25$ & \textbf{61.4} & \textbf{65.9} & \textbf{68.3} & \textbf{72.1} & \textbf{51.3} & \textbf{52.4} & \textbf{64.7} & \textbf{69.1} & \textbf{50.6} & \textbf{56.4} & \textbf{70.1} & \textbf{73.8} \\
\midrule
\multirow{2}[2]{*}{bce-embedding-base-v1} & -     & 59.1  & 63.8  & -     & -     & 42.0  & 44.0  & -     & -     & -     & -     & -     & - \\
      & QAE\textsubscript{hyb}, $\alpha=0.3,\;\beta=0.5$ & \textbf{66.8} & \textbf{71.1} & -     & -     & \textbf{49.7} & \textbf{51.0} & -     & -     & -     & -     & -     & - \\
\midrule
\multirow{2}[2]{*}{text2vec-base-multilingual} & -     & 38.7  & 43.6  & 27.8  & 31.9  & 9.7   & 10.6  & 33.7  & 38.6  & 15.6  & 19.9  & 12.6  & 15.5 \\
      & QAE\textsubscript{emb}, $\alpha=0.75$ & \textbf{55.4} & \textbf{59.9} & \textbf{51.5} & \textbf{55.4} & \textbf{32.1} & \textbf{34.1} & \textbf{49.3} & \textbf{54.4} & \textbf{36.2} & \textbf{41.5} & \textbf{47.1} & \textbf{52.1} \\
\midrule
\multicolumn{14}{c}{\textbf{Ablation}} \\
\midrule
\multirow{4}[2]{*}{bge-m3} & -     & 74.4  & 78.7  & 77.8  & 80.9  & 47.5  & 48.6  & 73.5  & 77.4  & 58.6  & 64.4  & 78.7  & 81.4 \\
      & QAE\textsubscript{emb}, $\alpha=0.3$ & 76.4  & 80.5  & 80.1  & 82.9  & 51.3  & 52.4  & 75    & 78.5  & 61    & 66    & 79    & 82 \\
      & QAE\textsubscript{txt}, $\beta=1.5$ & 77.2  & 81    & 80.2  & 83.1  & 51.4  & 52.5  & 76.9  & 80.3  & 62.7  & 67.8  & 80    & 82.8 \\
      & QAE\textsubscript{hyb}, $\alpha=0.15,\;\beta=1.5$ & 77.4  & 81.1  & 80.6  & 83.4  & 51.7  & 52.7  & 76.5  & 80    & 61.7  & 66.8  & 79.8  & 82.7 \\
\midrule
\multirow{4}[2]{*}{multilingual-e5-small} & -     & 71    & 75.1  & 74.1  & 77.4  & 44.6  & 46.0  & 66.8  & 70.4  & 52.5  & 57.8  & 72.9  & 76.5 \\
      & QAE\textsubscript{emb}, $\alpha=0.45$ & 74.7  & 78.5  & 77    & 79.8  & 50.8  & 51.9  & 73.6  & 77.1  & 58    & 63.2  & 77.1  & 80.1 \\
      & QAE\textsubscript{txt}, $\beta=1.0$ & 73.2  & 77.2  & 79.2  & 81.9  & 49.1  & 50.5  & 70.6  & 74.7  & 58.7  & 63.7  & 77.1  & 80.2 \\
      & QAE\textsubscript{hyb}, $\alpha=0.3,\;\beta=0.5$ & 74.6  & 78.5  & 78.9  & 81.6  & 50.6  & 51.6  & 74.2  & 77.9  & 59.6  & 64.6  & 77.5  & 80.4 \\
\midrule
\multirow{4}[2]{*}{multilingual-e5-base} & -     & 74.8  & 78.1  & 72.3  & 76    & 47.0  & 48.2  & 71.2  & 75.1  & 57.8  & 62.9  & 72.6  & 75.8 \\
      & QAE\textsubscript{emb}, $\alpha=0.3$ & 77.6  & 81.3  & 77.2  & 80.3  & 51.2  & 52.3  & 76.7  & 80    & 61.5  & 66.5  & 77.7  & 80.5 \\
      & QAE\textsubscript{txt}, $\beta=0.75$ & 74.7  & 78.8  & 76.4  & 79.7  & 50.6  & 51.8  & 72.3  & 76.4  & 62.5  & 67.8  & 77    & 79.9 \\
      & QAE\textsubscript{hyb}, $\alpha=0.3,\;\beta=0.5$ & 76.6  & 80.4  & 77.4  & 80.6  & 51.5  & 52.5  & 75.4  & 79.1  & 62.7  & 67.7  & 77.2  & 80.2 \\
\midrule
\multirow{4}[2]{*}{multilingual-e5-large} & -     & 73.9  & 77.8  & 76.7  & 80.2  & 46.9  & 48.3  & 70.6  & 74.5  & 53    & 59.2  & 73.9  & 77.4 \\
      & QAE\textsubscript{emb}, $\alpha=0.45$ & 77.9  & 81.4  & 79.8  & 83    & 51.9  & 52.9  & 77    & 80    & 58.3  & 63.5  & 78.1  & 81.2 \\
      & QAE\textsubscript{txt}, $\beta=1.5$ & 75.6  & 79.2  & 80.9  & 84.1  & 51.0  & 52.3  & 76    & 79.4  & 60    & 65.5  & 77    & 80.2 \\
      & QAE\textsubscript{hyb}, $\alpha=0.15,\;\beta=1.25$ & 77.1  & 80.6  & 82.2  & 85.1  & 51.5  & 52.7  & 77.4  & 80.9  & 60.6  & 65.9  & 77.7  & 81 \\
\midrule
\multirow{4}[2]{*}{gte-multilingual-base} & -     & 65.5  & 70.4  & 73.4  & 76.8  & 45.3  & 46.8  & 63    & 67.4  & 52.4  & 58.6  & 66    & 69.6 \\
      & QAE\textsubscript{emb}, $\alpha=0.45$ & 69.1  & 73.6  & 76.9  & 79.6  & 50.5  & 51.7  & 67.1  & 71.2  & 53    & 58.5  & 72.4  & 75.8 \\
      & QAE\textsubscript{txt}, $\beta=1.5$ & 75.7  & 79.7  & 75.9  & 78.9  & 48.8  & 50.5  & 66.1  & 70.7  & 56.8  & 62.5  & 72.8  & 76.5 \\
      & QAE\textsubscript{hyb}, $\alpha=0.15,\;\beta=1.5$ & 75.5  & 79.5  & 76.2  & 79.1  & 49.4  & 51.0  & 66.9  & 71.6  & 56    & 61.9  & 74    & 77.6 \\
\midrule
\multirow{4}[2]{*}{mcontriever} & -     & 32.9  & 36.7  & 40.3  & 44.7  & 39.2  & 41.6  & 35    & 39.3  & 27    & 31.8  & 49.5  & 54.4 \\
      & QAE\textsubscript{emb}, $\alpha=0.6$ & 58.8  & 64.2  & 67.2  & 71.3  & 51.0  & 52.1  & 62.7  & 66.9  & 50.4  & 55.7  & 69.7  & 73.2 \\
      & QAE\textsubscript{txt}, $\beta=1.5$ & 49.2  & 54.2  & 59.5  & 63.9  & 46.2  & 48.2  & 54.1  & 58.9  & 45.1  & 50.2  & 60.8  & 64.6 \\
      & QAE\textsubscript{hyb}, $\alpha=0.45,\;\beta=1.25$ & 61.4  & 65.9  & 68.3  & 72.1  & 51.3  & 52.4  & 64.7  & 69.1  & 50.6  & 56.4  & 70.1  & 73.8 \\
\midrule
\multirow{4}[2]{*}{bce-embedding-base-v1} & -     & 59.1  & 63.8  & -     & -     & 42.0  & 44.0  & -     & -     & -     & -     & -     & - \\
      & QAE\textsubscript{emb}, $\alpha=0.45$ & 66.8  & 71.3  & -     & -     & 49.3  & 50.6  & -     & -     & -     & -     & -     & - \\
      & QAE\textsubscript{txt}, $\beta=1.5$ & 64.3  & 68.7  & -     & -     & 47.6  & 49.2  & -     & -     & -     & -     & -     & - \\
      & QAE\textsubscript{hyb}, $\alpha=0.3,\;\beta=0.5$ & 66.8  & 71.1  & -     & -     & 49.7  & 51.0  & -     & -     & -     & -     & -     & - \\
\midrule
\multirow{4}[2]{*}{text2vec-base-multilingual} & -     & 38.7  & 43.6  & 27.8  & 31.9  & 9.7   & 10.6  & 33.7  & 38.6  & 15.6  & 19.9  & 12.6  & 15.5 \\
      & QAE\textsubscript{emb}, $\alpha=0.75$ & 55.4  & 59.9  & 51.5  & 55.4  & 32.1  & 34.1  & 49.3  & 54.4  & 36.2  & 41.5  & 47.1  & 52.1 \\
      & QAE\textsubscript{txt}, $\beta=1.5$ & 46.2  & 50.8  & 35.5  & 39.4  & 12.3  & 13.6  & 43.3  & 47.7  & 21.3  & 25.2  & 18.8  & 22.1 \\
      & QAE\textsubscript{hyb}, $\alpha=0.75,\;\beta=0.5$ & 55.1  & 59.7  & 50.4  & 54.4  & 33.4  & 35.1  & 49.1  & 53.8  & 34.5  & 39.6  & 47.1  & 51.5 \\
\bottomrule
\end{tabular}%

    }

  \caption{\textcolor{black}{Comprehensive retrieval performance on the latest datasets FIGNEWS  and CRUD-RAG (\textbf{Top-k = 10}). Higher is better, with the best one bolded. Hyperparameters including QAEncoder variants and weight terms $\alpha,\;\beta$ are optimized simultaneously for six latest datasets. `-' denotes default or null values.}}
  \label{full_latest}%
\end{table*}%

\begin{table*}[htbp]
  \centering
  \vspace{10pt}
  \resizebox{\textwidth}{!}{

    \begin{tabular}{cccccccccccccc}
    \toprule
    \multirow{2}[2]{*}{Model} & \multirow{2}[2]{*}{Method} & \multicolumn{2}{c}{FIGNEWS(English)} & \multicolumn{2}{c}{FIGNEWS(Arabic)} & \multicolumn{2}{c}{CRUD-RAG(Chinese)} & \multicolumn{2}{c}{FIGNEWS(French)} & \multicolumn{2}{c}{FIGNEWS(Hindi)} & \multicolumn{2}{c}{FIGNEWS(Hebrew)} \\
          &       & \multicolumn{1}{l}{MRR@10} & \multicolumn{1}{l}{NDCG@10} & \multicolumn{1}{l}{MRR@10} & \multicolumn{1}{l}{NDCG@10} & \multicolumn{1}{l}{MRR@10} & \multicolumn{1}{l}{NDCG@10} & \multicolumn{1}{l}{MRR@10} & \multicolumn{1}{l}{NDCG@10} & \multicolumn{1}{l}{MRR@10} & \multicolumn{1}{l}{NDCG@10} & \multicolumn{1}{l}{MRR@10} & \multicolumn{1}{l}{NDCG@10} \\
    \midrule
    \multirow{4}[2]{*}{bg3-m3} & QAE\textsubscript{emb}, $\alpha=0.3$ & 76.4  & 80.5  & 80.1  & 82.9  & 88.8  & 90.7  & 75    & 78.5  & 61    & 66    & 79    & 82 \\
          & QAE\textsubscript{txt}, $\beta=1.5$ & 77.2  & 81    & 80.2  & 83.1  & 89    & 90.9  & \textbf{76.9} & \textbf{80.3} & \textbf{62.7} & \textbf{67.8} & \textbf{80} & \textbf{82.8} \\
          & QAE\textsubscript{hyb}, $\alpha=0.15,\;\beta=1.5$ & \textbf{77.4} & \textbf{81.1} & \textbf{80.6} & \textbf{83.4} & \textbf{89.4} & \textbf{91.2} & 76.5  & 80    & 61.7  & 66.8  & 79.8  & 82.7 \\
          & QA\textsubscript{naive}, n=10 & 76.9  & 79.9  & 77.1  & 79.7  & 86    & 88    & 71.9  & 74.4  & 62.3  & 66.1  & 68.1  & 71.7 \\
    \midrule
    \multirow{4}[2]{*}{multilingual-e5-large} & QAE\textsubscript{emb}, $\alpha=0.45$ & \textbf{77.9} & \textbf{81.4} & 79.8  & 83    & \textbf{89.8} & \textbf{91.5} & 77    & 80    & 58.3  & 63.5  & \textbf{78.1} & \textbf{81.2} \\
          & QAE\textsubscript{txt}, $\beta=1.5$ & 75.6  & 79.2  & 80.9  & 84.1  & 88.3  & 90.5  & 76    & 79.4  & 60    & 65.5  & 77    & 80.2 \\
          & QAE\textsubscript{hyb}, $\alpha=0.15,\;\beta=1.25$ & 77.1  & 80.6  & \textbf{82.2} & \textbf{85.1} & 89.1  & 91.2  & \textbf{77.4} & \textbf{80.9} & 60.6  & \textbf{65.9} & 77.7  & 81 \\
          & QA\textsubscript{naive}, n=10 & 77.5  & 80.3  & 76.5  & 79.4  & 85.1  & 87.3  & 70.5  & 73.5  & \textbf{61.5} & 65.6  & 69.4  & 72.2 \\
\bottomrule
    \end{tabular}%

    }

  \caption{\textcolor{black}{Complete performance comparison of QAEncoder variants on latest datasets FIGNEWS and CRUD-RAG (Top-k = 10). Higher is better, with the best one bolded. Hyperparameters are optimized simultaneously across  the six latest datasets. $n$ indicates the number of predicted queries in QA\textsubscript{naive}.}}
    \label{full:ablation}%
\end{table*}%

\begin{table*}[htbp]
  \centering
  \vspace{10pt}
  \resizebox{\textwidth}{!}{

\begin{tabular}{cccccccccccccc}
\toprule
\multirow{2}[4]{*}{Model} & \multirow{2}[4]{*}{Method} & \multicolumn{2}{c}{FIGNEWS(English)} & \multicolumn{2}{c}{FIGNEWS(Arabic)} & \multicolumn{2}{c}{CRUD-RAG(Chinese)} & \multicolumn{2}{c}{FIGNEWS(French)} & \multicolumn{2}{c}{FIGNEWS(Hindi)} & \multicolumn{2}{c}{FIGNEWS(Hebrew)} \\
\cmidrule{3-14}      &       & \multicolumn{1}{l}{MRR@10} & \multicolumn{1}{l}{NDCG@10} & \multicolumn{1}{l}{MRR@10} & \multicolumn{1}{l}{NDCG@10} & \multicolumn{1}{l}{MRR@10} & \multicolumn{1}{l}{NDCG@10} & \multicolumn{1}{l}{MRR@10} & \multicolumn{1}{l}{NDCG@10} & \multicolumn{1}{l}{MRR@10} & \multicolumn{1}{l}{NDCG@10} & \multicolumn{1}{l}{MRR@10} & \multicolumn{1}{l}{NDCG@10} \\
\midrule
\multirow{8}[2]{*}{mcontriever} & -     & 32.9  & 36.7  & 40.3  & 44.7  & 39.2  & 41.6  & 35    & 39.3  & 27    & 31.8  & 49.5  & 54.4 \\
      & QAE\textsubscript{hyb}, $\alpha=0.45,\;\beta=1.25$ & 61.4  & 65.9  & 68.3  & 72.1  & 51.3  & 52.4  & 64.7  & 69.1  & 50.6  & 56.4  & 70.1  & 73.8 \\
      & MS$^{\dagger}$    & 66.1  & 70.6  & 70.2  & 73.7  & 46.5  & 47.8  & 66.4  & 70.3  & 48.7  & 53.9  & 69.4  & 72.9 \\
      & MS$^{\dagger}$ + QAE\textsubscript{hyb},  $\alpha=0.3,\;\beta=0.75$ & \textbf{72.3} & \textbf{76.8} & \textbf{77.3} & \textbf{80.5} & 51.2  & 52.4  & \textbf{74.4} & \textbf{78} & \textbf{59} & \textbf{64.1} & \textbf{76.5} & \textbf{79.8} \\
      & GPL$^{\dagger}$   & 68.3  & 73.6  & 72.1  & 75.18 & \textbf{52.8} & \textbf{55.9} & 70.8  & 74.3  & 52.6  & 55.9  & 73.9  & 77.1 \\
      & QA-RAG$^{\ddagger}$ & 31.1  & 34.4  & 42.3  & 46.8  & 43.8  & 45.8  & 31.8  & 35.5  & 29.1  & 33.9  & 49.7  & 55.0 \\
      & Query2Doc$^{\ddagger}$ & 25.1  & 29.0  & 34.5  & 38.9  & 35.7  & 37.2  & 26.0  & 29.1  & 14.2  & 17.2  & 39.9  & 45.0 \\
      & HyDE$^{\ddagger}$  & 25    & 27.9  & 35.7  & 41.9  & 36.7  & 38.7  & 25.8  & 28.9  & 11.8  & 14.9  & 42.1  & 47.8 \\
\midrule
\multirow{8}[2]{*}{multilingual-e5-large} & -     & 73.9  & 77.8  & 76.7  & 80.2  & 46.9  & 48.3  & 70.6  & 74.5  & 53    & 59.2  & 73.9  & 77.4 \\
      & QAE\textsubscript{hyb}, $\alpha=0.15,\;\beta=1.25$ & \textbf{77.1} & \textbf{80.6} & \textbf{82.2} & \textbf{85.1} & 51.5  & 52.7  & \textbf{77.4} & \textbf{80.9} & \textbf{60.6} & \textbf{65.9} & 77.7  & 81 \\
      & INS$^{\dagger}$   & 67    & 71.4  & 75    & 78.2  & 43.7  & 45.2  & 66    & 70.9  & 48.2  & 53.8  & 69.6  & 73.3 \\
      & INS$^{\dagger}$ + QAE\textsubscript{hyb},  $\alpha=0.15,\;\beta=1.5$ & 75.6  & 79.8  & 80.8  & 83.7  & 51.4  & 52.4  & 75.8  & 79.5  & 59.3  & 65    & \textbf{79.4} & \textbf{82.3} \\
      & GPL$^{\dagger}$   & 75.2  & 78.9  & 79.4  & 82.3  & \textbf{53.6} & \textbf{56.3} & 73.7  & 76.8  & 57.5  & 62.8  & 74.8  & 78.3 \\
      & QA-RAG$^{\ddagger}$ & 73.3  & 76.5  & 72.8  & 75.6  & 45.8  & 46.5  & 67.2  & 70.2  & 55.5  & 61.3  & 77.7  & 80.4 \\
      & Query2Doc$^{\ddagger}$ & 63.4  & 68.2  & 66.3  & 72.8  & 42.0  & 43.1  & 60.4  & 65.1  & 48.4  & 53.3  & 62.1  & 66.0 \\
      & HyDE$^{\ddagger}$  & 63.6  & 68.3  & 68.3  & 74.1  & 42.3  & 43.6  & 58.4  & 63.2  & 45.6  & 51.3  & 65.3  & 69.6 \\
\bottomrule
\end{tabular}%

    }

  \caption{\textcolor{black}{The table illustrates a comprehensive performance comparison of QAEncoder against training-based and document-centric methods on the latest datasets: FIGNEWS and CRUD-RAG (Top-k = 10). Higher is better, with the best one bolded. Hyperparameters $\alpha,\;\beta$ are optimized simultaneously across the six latest datasets. 
   $\dagger$ indicates the training-based methods. MS$^{\dagger}$ represents fine-tuning on MSMARCO, i.e. the mcontriever-msmarco model; INS$^{\dagger}$ represents instruction-tuning, i.e. the multilingual-e5-large-instruct model\citep{wang2023improving}. $\ddagger$ indicates the document-centric methods.}}

    \label{full:dis}
\end{table*}%

\begin{table*}[htbp]
  \centering
  \vspace{10pt}
  \resizebox{\textwidth}{!}{

\begin{tabular}{cccccccc}
\toprule
\multirow{2}[4]{*}{Model} & \multirow{2}[4]{*}{Param} & \multicolumn{2}{c}{FIGNEWS(English)} & \multicolumn{2}{c}{FIGNEWS(Arabic)} & \multicolumn{2}{c}{CRUD-RAG(Chinese)} \\
\cmidrule{3-8}      &       & MRR@10 & NDCG@10 & MRR@10 & NDCG@10 & MRR@10 & NDCG@10 \\
\midrule
\multirow{2}[2]{*}{jina-embeddings-v2-small-en} & -     & 65.4  & 69.9  & -     & -     & -     & - \\
      & QAE\textsubscript{hyb}, $\alpha=0.15,\;\beta=1.25$ & \textbf{72.6} & \textbf{76.7} & -     & -     & \textbf{-} & \textbf{-} \\
\midrule
\multirow{2}[2]{*}{jina-embeddings-v2-base-zh} & -     & -     & -     & -     & -     & 41.0  & 43.1 \\
      & QAE\textsubscript{hyb}, $\alpha=0.75,\;\beta=0.75$ & \textbf{-} & \textbf{-} & \textbf{-} & \textbf{-} & \textbf{47.5} & \textbf{48.5} \\
\midrule
\multirow{2}[2]{*}{jina-embeddings-v2-base-en} & -     & 65.3  & 69.4  & -     & -     & -     & - \\
      & QAE\textsubscript{hyb}, $\alpha=0.3,\;\beta=0.5$ & \textbf{72.4} & \textbf{76.5} & \textbf{-} & \textbf{-} & \textbf{-} & \textbf{-} \\
\midrule
\multirow{2}[2]{*}{gte-base-en-v1.5} & -     & 65.6  & 70.2  & -     & -     & -     & - \\
      & QAE\textsubscript{hyb}, $\alpha=0.15,\;\beta=1.5$ & \textbf{71.3} & \textbf{75.6} & \textbf{-} & \textbf{-} & \textbf{-} & \textbf{-} \\
\midrule
\multirow{2}[2]{*}{contriever} & -     & 49.6  & 54.5  & -     & -     & -     & - \\
      & QAE\textsubscript{hyb}, $\alpha=0.45,\;\beta=1.25$ & \textbf{70.7} & \textbf{74.7} & \textbf{-} & \textbf{-} & \textbf{-} & \textbf{-} \\
\midrule
\multirow{2}[2]{*}{bge-large-zh-v1.5} & -     & -     & -     & -     & -     & 42.0  & 43.7 \\
      & QAE\textsubscript{hyb}, $\alpha=0.45,\;\beta=1.25$ & \textbf{-} & \textbf{-} & -     & -     & \textbf{48.6} & \textbf{49.6} \\
\midrule
\multirow{2}[2]{*}{bge-large-zh} & -     & -     & -     & -     & -     & 40.6  & 42.5 \\
      & QAE\textsubscript{hyb}, $\alpha=0.45,\;\beta=1.5$ & \textbf{-} & \textbf{-} & \textbf{-} & \textbf{-} & \textbf{48.4} & \textbf{49.4} \\
\midrule
\multirow{2}[2]{*}{bge-large-en-v1.5} & -     & 66.4  & 71    & -     & -     & -     & - \\
      & QAE\textsubscript{hyb}, $\alpha=0.15,\;\beta=1.5$ & \textbf{74.3} & \textbf{78.2} & \textbf{-} & \textbf{-} & \textbf{-} & \textbf{-} \\
\midrule
\multirow{2}[2]{*}{bge-large-en} & -     & 61.9  & 66.4  & -     & -     & -     & - \\
      & QAE\textsubscript{hyb}, $\alpha=0.3,\;\beta=0.75$ & \textbf{71.3} & \textbf{75.4} & \textbf{-} & \textbf{-} & \textbf{-} & \textbf{-} \\
\midrule
\multirow{2}[2]{*}{bge-base-zh-v1.5} & -     & -     & -     & -     & -     & 43.4  & 44.9 \\
      & QAE\textsubscript{hyb}, $\alpha=0.6,\;\beta=1.0$ & \textbf{-} & \textbf{-} & \textbf{-} & \textbf{-} & \textbf{48.6} & \textbf{49.5} \\
\midrule
\multirow{2}[2]{*}{bge-base-zh} & -     & -     & -     & -     & -     & 40.3  & 41.9 \\
      & QAE\textsubscript{hyb}, $\alpha=0.6,\;\beta=1.5$ & \textbf{-} & \textbf{-} & \textbf{-} & \textbf{-} & \textbf{48.5} & \textbf{49.5} \\
\midrule
\multirow{2}[2]{*}{bge-base-en-v1.5} & -     & 66.4  & 70.5  & -     & -     & -     & - \\
      & QAE\textsubscript{txt}, $\beta=1.0$ & \textbf{74.1} & \textbf{77.5} & -     & -     & \textbf{-} & \textbf{-} \\
\midrule
\multirow{2}[2]{*}{bge-base-en} & -     & 64.9  & 68.9  & -     & -     & -     & - \\
      & QAE\textsubscript{hyb}, $\alpha=0.3,\;\beta=1.5$ & \textbf{71.7} & \textbf{75.9} & \textbf{-} & \textbf{-} & \textbf{-} & \textbf{-} \\
\midrule
\multirow{2}[1]{*}{bce-embedding-base-v1} & -     & 59.1  & 63.8  & -     & -     & 42.0  & 44.0 \\
      & QAE\textsubscript{hyb}, $\alpha=0.3,\;\beta=0.5$ & \textbf{66.8} & \textbf{71.1} & \textbf{-} & \textbf{-} & \textbf{47.0} & \textbf{48.3} \\
\midrule
\multirow{2}[1]{*}{retriever-multi-base-256} & -     & 33.8  & 37.4  & -     & -     & -  & - \\
      & QAE\textsubscript{emb}, $\alpha=0.75$ & \textbf{58.4} & \textbf{62.3} & \textbf{-} & \textbf{-} & \textbf{-} & \textbf{-} \\
\midrule
\multirow{2}[1]{*}{quora-distilbert-base} & -     & 29.8  & 33.2  & -     & -     & -  & - \\
      & QAE\textsubscript{emb}, $\alpha=0.75$ & \textbf{52.8} & \textbf{57.1} & \textbf{-} & \textbf{-} & \textbf{-} & \textbf{-} \\
      \bottomrule
\end{tabular}%
    }

  \caption{\textcolor{black}{
  Retrieval performance of \textbf{monolingual and bilingual embedding models} on the latest datasets FIGNEWS(English) and CRUD-RAG(Chinese). Higher is better, with the best one bolded. `-' denotes default or null values due to mismatch between the language and model.
  }}
    \label{cn_en}%
\end{table*}%

\end{document}